\documentclass[10pt,twocolumn,letterpaper]{article}

\usepackage{titling}
\usepackage{cvpr}
\usepackage{times}
\usepackage{epsfig}
\usepackage{graphicx}
\usepackage{amsmath}
\usepackage{amssymb}
\usepackage{amsthm}
\usepackage{multibib}

\usepackage{bbding}
\usepackage{tabularx}
\usepackage{booktabs}
\usepackage{multirow}
\usepackage{newtxtt}
\usepackage{etoolbox,siunitx}
\robustify\bfseries
\usepackage{color}
\definecolor{mygreen}{rgb}{0.2, 0.7, 0.1}
\usepackage[pagebackref=true,breaklinks=true,letterpaper=true,colorlinks,citecolor={mygreen},bookmarks=false]{hyperref}

\usepackage[capitalize]{cleveref}
\crefname{section}{Sec.}{Section}

\usepackage[format=plain,labelformat=simple,labelsep=period,font=small,skip=4pt,compatibility=false]{caption}
\usepackage[font=scriptsize,skip=2pt,subrefformat=parens]{subcaption}

\usepackage{pifont} 
\newcommand{\cmark}{\ding{51}}%
\newcommand{\xmark}{\ding{55}}%

\usepackage{xspace}

\newcommand*{\myparagraph}[1]{\smallskip\noindent\textbf{#1}\hspace{0.5em}}

\cvprfinalcopy 

\def\cvprPaperID{3185} 

\ifcvprfinal\fi

\newcommand\blfootnote[1]{%
  \begingroup
  \renewcommand\thefootnote{}\footnote{#1}%
  \addtocounter{footnote}{-1}%
  \endgroup
}

\usepackage{fancyhdr}
\usepackage{setspace}

\newcites{supp}{References}

\begin{document}

\title{Single-Stage Semantic Segmentation from Image Labels}

\author{Nikita Araslanov \hspace{1cm} Stefan Roth\\
Department of Computer Science, TU Darmstadt}

\maketitle
\thispagestyle{fancy}

\begin{abstract}

Recent years have seen a rapid growth in new approaches improving the accuracy of semantic segmentation in a weakly supervised setting, \ie with only image-level labels available for training.
However, this has come at the cost of increased model complexity and sophisticated multi-stage training procedures.
This is in contrast to earlier work that used only a \emph{single stage} -- training one segmentation network on image labels --
which was abandoned due to inferior segmentation accuracy.
In this work, we first define three desirable properties of a weakly supervised method: \emph{local consistency}, \emph{semantic fidelity}, and \emph{completeness}.
Using these properties as guidelines, we then develop a segmentation-based network model and a self-supervised training scheme to train for semantic masks from image-level annotations in a \emph{single stage}.
We show that despite its simplicity, our method achieves results that are competitive with significantly more complex pipelines, substantially outperforming earlier single-stage methods.

\blfootnote{
Code is available at \href{https://github.com/visinf/1-stage-wseg}{https://github.com/visinf/1-stage-wseg}.
}
\end{abstract}

\section{Introduction}

Many applications of scene understanding require some form of semantic localisation with pixel-level precision, hence semantic segmentation has enjoyed enormous popularity.
Despite the successes of supervised learning approaches \cite{deeplabv3plus2018,long2015fully}, their general applicability remains limited due to their reliance on pixel-level annotations.
We thus consider the task of learning semantic segmentation from image-level annotations alone, aiming to develop a \emph{practical} approach.
This problem setup is especially challenging compared to other weakly supervised scenarios that assume available localisation cues, such as bounding boxes, scribbles, and points \cite{bearman2016s,dai2015boxsup,KhorevaBH0S17,lin2016scribblesup,tang2018regularized}. 

\begin{figure}[t]%
    \centering
    \includegraphics[width=\linewidth]{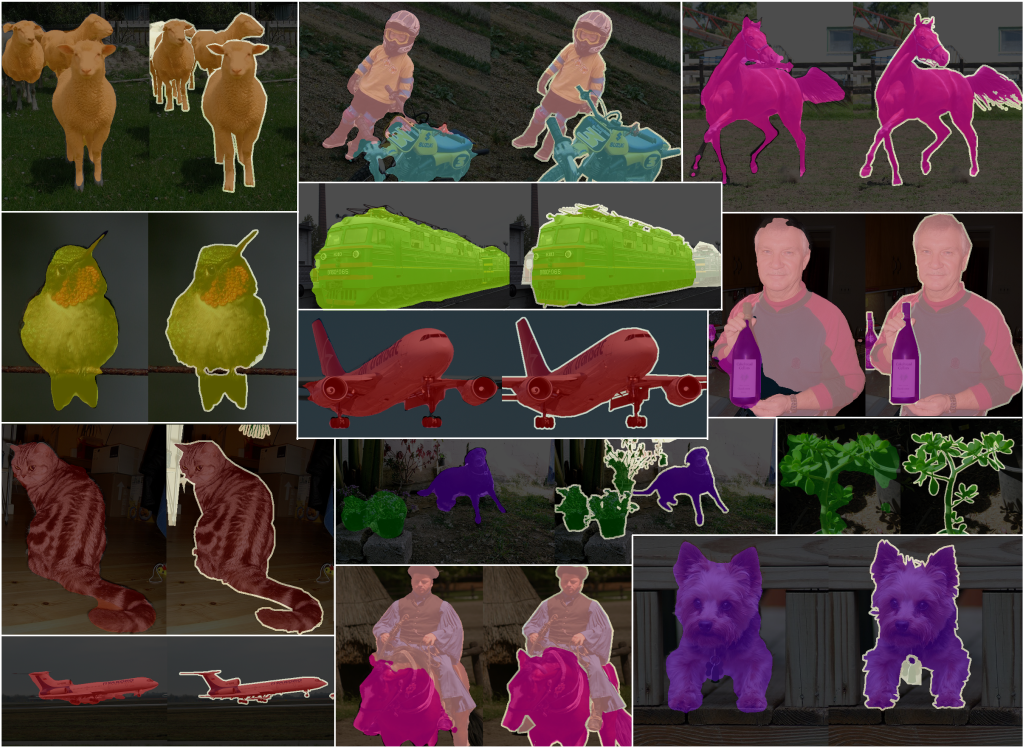}
    \caption{Our segmentation model can produce high-quality semantic masks (\textit{left}) approaching the ground truth (\textit{right}) in a single stage of training while using only image-level annotation.
    No post-processing was applied to these examples.}
    \label{fig:teaser}%
    \vspace{-0.5em}
\end{figure}

Attention mechanisms, such Class Activation Maps (CAM)~\cite{ZhouKLOT16}, offer a partial solution: 
they localise the most discriminative regions in the image using only a pre-trained classification network.
Such masks, however, are quite coarse -- they violate object boundaries, tend to be incomplete for large-scale objects and imprecise for small ones.
This is not surprising, since attention maps were not designed for segmentation in the first place. 
Nevertheless, most methods for weakly supervised segmentation from image labels adopt attention maps (\eg, CAMs) as initial \emph{seeds} for further refinement.
Yet the remarkable progress these methods have achieved -- currently reaching more than 80\% of fully supervised accuracy~\cite{AhnK18,WeiXSJFH18} -- has come along with increased model and training complexity.
While early methods consisted of a single stage, \ie~training one network \cite{HongOLH16,PapandreouCMY15,PathakSLD14,PinheiroC15}, they were soon superseded by more advanced pipelines, employing multiple models, training cycles, and
off-the-shelf saliency methods \cite{HuangWWLW18,WeiFLCZY17,WeiXSJFH18,Zeng_2019_ICCV}.

\begin{figure}[t]
\def\svgwidth{\linewidth}
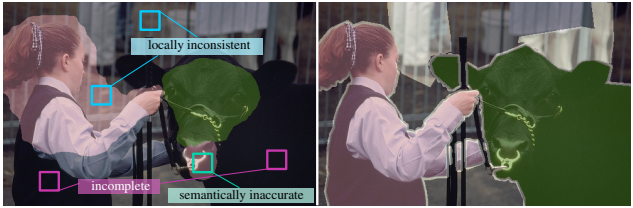
\caption{\textbf{Typical shortcomings of attention maps} \emph{(left)} \vs pixel-level annotations \emph{(right)}.
Our observations of failure modes in attention maps lead to three defining properties of segmentation-aware training from image labels, which we incorporate into our model: \emph{semantic fidelity}, \emph{local consistency}, and \emph{completeness}.}
\label{fig:cam_problems}
\vspace{-0.5em}
\end{figure}

In this work, we develop an effective single-stage approach for weakly supervised semantic segmentation that streamlines previous multi-stage attempts, and uses neither saliency estimation nor additional data.
Our key insight is to enable \emph{segmentation-aware} training for classification.
Consider \cref{fig:cam_problems} depicting typical limitations of attention maps:
\textit{(i)} two areas in local proximity with similar appearance may be assigned different classes, \ie the semantic labelling may be \emph{locally} inconsistent;
\textit{(ii)} attention maps tend to be incomplete in terms of covering the whole extent of the object;
\textit{(iii)} while the area of the attention maps dominates for the correct object class, parts of the map may still be mislabelled (\ie are semantically inaccurate).
These observations lead us to define three properties a \emph{segmentation-aware} training should encompass:
\emph{(a)} \emph{local consistency} implies that neighbouring pixels with similar appearance share the same label;
\emph{(b)} \emph{semantic fidelity} is exhibited by models producing segmentation masks that allow for reliable classification decisions (\eg, with good generalisation);
 \emph{(c)} \emph{completeness} means that our model identifies all visible class occurrences in the image.
Note that since classification requires only sufficient evidence, CAMs neither ensure \emph{completeness} nor \emph{local consistency}.

Using these concepts as our guidelines, we design an approach that significantly outperforms CAMs in terms of segmentation accuracy.
First, we propose \emph{normalised Global Weighted Pooling}, a novel process for computing the classification scores, which enables concurrent training for the segmentation task.
Second, we encourage masks to heed appearance cues with \textit{Pixel-Adaptive Mask Refinement}.
These masks are supplied to our model as pseudo ground truth for self-supervised segmentation.
Third, to counter the compounding effect of inaccuracies in the pseudo mask annotation (a common problem of self-supervised methods), we introduce a \emph{Stochastic Gate} that mixes feature representations with varying receptive field sizes.
As our experiments demonstrate, the resulting single-stage model offers segmentation quality on par or outperforming the state of the art, while being simple to train and to use.

\pagestyle{plain}
\section{Related Work}

Methods for weakly supervised semantic segmentation have evolved rapidly from simple single-stage models to more complex ones, employing saliency estimation methods, additional data (\eg, videos), and fully supervised ``fine-tuning'' (\cf \cref{sec:stages}).

\myparagraph{Single-stage methods.}
Following \cite{PathakSLD14}, Pinheiro \& Collobert \cite{PinheiroC15} used a Multiple Instance Learning (MIL) formulation, but applied a \texttt{LogSumExp}-aggregation of the pixel-level predictions in the output layer to produce the class scores, and refined the segments with image-level priors.
Papandreou \etal \cite{PapandreouCMY15} took an expectation-maximisation (EM) approach, where masks are inferred from intermediate predictions and used as pseudo ground truth. 
\cite{RoyT17} combines top-down attention masks with bottom-up segmentation cues in an end-to-end model using CRF-RNN \cite{Zheng15}.
The attention-based model \cite{HongOLH16} allows for joint classification and segmentation training in a cross-domain setting. 
Despite their simplicity, single-stage models have fallen out of favour, owing to their inferior segmentation accuracy.

\myparagraph{Seed and expand.}
Kolesnikov \& Lampert \cite{KolesnikovL16} introduced the idea of expanding high-precision localisation cues, such as CAMs \cite{ZhouKLOT16}, to align with segment boundaries.
In this framework, a segmentation network can be trained end-to-end, but the localisation cues are pre-computed from a standalone classification network.
Consequently, higher-quality localisation \cite{li2018tell} can further improve the segmentation accuracy.
Following the seed-and-expand principle, Huang \etal \cite{HuangWWLW18} employed a seeded region growing algorithm \cite{AdamsB94} to encourage larger coverage of the initial localisation seeds. 

\myparagraph{Erasing.}
One common observation with CAMs \cite{ZhouKLOT16} is their tendency to identify only the most discriminative class evidence.
Wei \etal \cite{WeiFLCZY17} explored the idea of ``erasing'' these high-confidence areas from the images and re-training the network for classification using the left-over regions to mine for additional cues.
Similarly, SeeNet \cite{HouJWC18} implements erasing using two decoder branches aided by saliency \cite{HouCHBTT19};
the first branch removes the peak CAM-response and feeds into the second.
To avoid re-training or modifying the decoder structure, Chaudhry \etal \cite{ChaudhryDT17} iteratively applied an off-the-shelf saliency detector \cite{LiuH16} to a progressively erased image in order to accumulate the foreground mask.

\myparagraph{Multiple training rounds.}
Another line of work trains a chain of segmentation networks, each learning the predictions of its predecessor \cite{KhorevaBH0S17}.
Wei \etal \cite{WeiLCSCFZY17} sequentially trained three networks in increasing order of task difficulty.
Following Khoreva \etal \cite{KhorevaBH0S17}, Jing \etal \cite{JingCT20} used multiple training rounds and refine intermediate results with GrabCut \cite{rother2004grabcut}, yet without the use of bounding-box annotations.
Similarly, Wang \etal \cite{wang2018weakly} iteratively train a network with the seed-and-expand principle, refine the intermediate results with saliency maps \cite{WangJYCHZ17}, and provide them as supervision to the segmentation network.

\myparagraph{Additional data.}
Hong \etal \cite{HongYKLH17} sought additional data in videos for training a class-agnostic decoder:
at inference time the class-specific attention map has to pass through the decoder individually.
More recently, Lee \etal \cite{Lee_2019_ICCV} aggregated additional attention maps from videos by merging the detected masks from consecutive frames via warping.

\myparagraph{Saliency and further refinement.}
Zeng \etal \cite{Zeng_2019_ICCV} showed that joint training with saliency ground truth significantly improves the mask accuracy.
Fan \etal \cite{fan2018associating} combined the saliency detector of \cite{fan2019s4net} with attention maps to partition the features within each detection window into a segment.
To increase the recall of attention maps, Wei \etal \cite{WeiXSJFH18} added multiple dilation rates in the last layer.
Towards the same goal, Lee \etal \cite{LeeKLLY19} stochastically selected hidden units and supplied the improved initial seeds to DSRG \cite{HuangWWLW18}.

\myparagraph{Image-level labels only.}
In this work we adhere to the early practice of relying only on image-label annotation.
Following this setup, Ahn \& Kwak \cite{AhnK18} modeled the pixel-level affinity distance from initial CAMs and employed a random walk to propagate individual class labels at the pixel level.
Instance-aware affinity encoding provides additional benefits \cite{AhnCK19}.
Both methods require training a standalone segmentation network on the masks for the final result -- a common practice (\eg, \cite{HuangWWLW18,WeiFLCZY17,WeiLCSCFZY17,WeiXSJFH18}).
Shimoda \& Yanai \cite{Shimoda_2019_ICCV} proposed to post-refine these masks with a cascade of three additional ``difference detection'' modules.

In contrast to these works, we develop a competitive alternative with practicality in mind: \emph{a single network for weakly supervised segmentation, trained in one cycle.}

\section{Model}

\subsection{Overview}

\begin{figure}[t]%
    \def\svgwidth{\linewidth}
    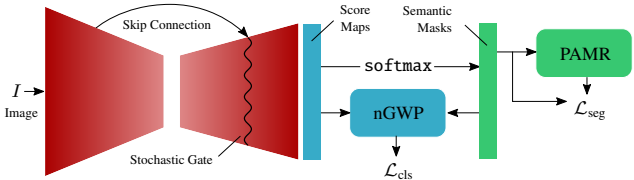
    \caption{\textbf{Architecture overview.} Our model shares the design of a segmentation network, but additionally makes use of \emph{normalised Global Weighted Pooling} (nGWP, \cref{sec:mask_size}) and \emph{Pixel-Adaptive Mask Refinement} (PAMR, \cref{sec:mask_refinement}) to enable self-supervised learning for segmentation from image labels.}%
    \label{fig:arch_overview}%
    \vspace{-0.5em}
\end{figure}

Our network model, illustrated in \cref{fig:arch_overview}, follows the established design of a fully convolutional segmentation network with a \texttt{softmax} output and skip connections \cite{long2015fully}.
This allows for a straightforward extension of any segmentation-based network architecture and exploiting pre-trained classification models for parameter pre-conditioning.
Inference requires only a single forward pass, analogous to fully supervised segmentation networks.    
By contrast, however, our model allows to learn for segmentation in a self-supervised fashion from image labels alone.

We propose three novel components relevant to our task:
\emph{(i)} a new class aggregation function, \emph{(ii)} a local mask refinement module, and \emph{(iii)} a stochastic gate.
The purpose of the class aggregation function is to leverage segmentation masks for classification decisions, \ie to provide \emph{semantic fidelity} as defined earlier.
To this end, we develop a \emph{normalised Global Weighted Pooling} (nGWP) that utilises pixel-level confidence predictions for relative weighting of the corresponding classification scores.
Additionally, we incorporate a \emph{focal mask penalty} into the classification scores to encourage \emph{completeness}.
We discuss these components in more detail in \cref{sec:mask_size}.
Next, in order to comply with \emph{local consistency}, we propose \emph{Pixel-Adaptive Mask Refinement} (PAMR), which revises the coarse mask predictions \wrt appearance cues.
The updated masks are further used as pseudo ground truth for segmentation, trained jointly along with the classification objective, as we explain in \cref{sec:mask_refinement}.
The refined masks produced by PAMR may still contain inaccuracies \wrt the ground truth, and self-supervised learning may further compound these errors via overfitting.
To alleviate these effects, we devise a \emph{Stochastic Gate} (SG) that combines a deep feature representation susceptible to this phenomenon with more robust, but less expressive shallow features in a stochastic way.
\cref{sec:stochasitc_gate} provides further detail.

\subsection{Classification scores}
\label{sec:mask_size}

\begin{figure}[t]
\subcaptionbox{\label{fig:GAP}}{
\def\svgwidth{\linewidth}
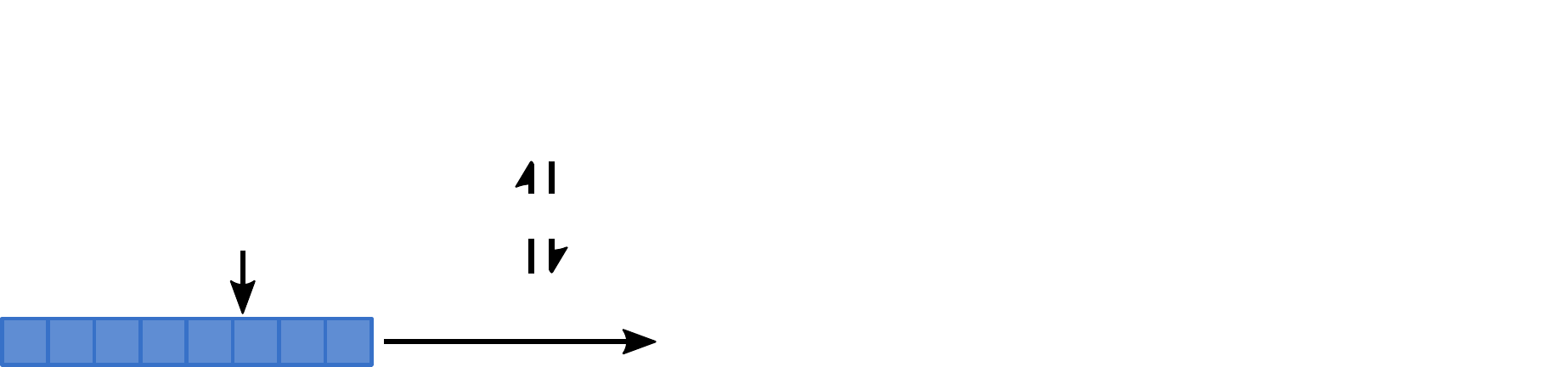
}\vspace{1.0em}
\subcaptionbox{\label{fig:GWP}}{
\def\svgwidth{\linewidth}
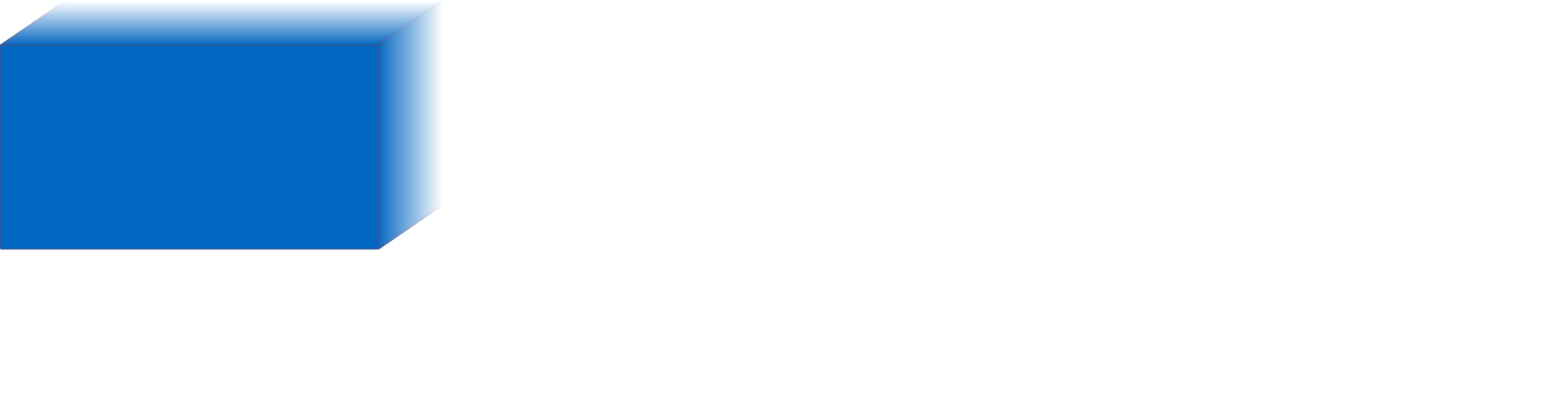
}

\caption{\textbf{The original GAP-CAM architecture} \subref{fig:GAP} and our \textbf{proposed modification, nGWP} \subref{fig:GWP}.
Our analysis of CAMs prompts us to devise an alternative aggregation mapping of class scores, nGWP, which allows to re-use the original classification loss, yet enables joint training for segmentation with substantial improvements in mask quality.
}
\vspace{-0.5em}
\end{figure}

\paragraph{CAMs.}
It is instructive to briefly review how the class score is normally computed with Global Average Pooling (GAP), since this analysis builds the premise for our aggregation mapping.
Let $x_{k,:,:}$ denote one of $K$ feature channels of size $h \times w$ preceding GAP, and $a_{c,:}$ be the parameter vector for class $c$ in the fully connected prediction layer.
The class score for class $c$ is then obtained as
\begin{equation}
\begin{aligned}
y^{\text{GAP}}_c = \frac{1}{h w} \sum_{k=1}^K a_{c,k} \sum_{i,j} x_{k,i,j}.
\end{aligned}
\label{eq:gap_score}
\end{equation}
Next, we can compute the Class Activation Mapping (CAM) \cite{ZhouKLOT16} for class $c$ as
\begin{equation}
\begin{aligned}
m^{\text{CAM}}_{c,:,:} = \max\bigg(0, \sum_{k=1}^K a_{c,k} x_{k,:,:} \bigg).
\end{aligned}
\label{eq:cam}
\end{equation}
\cref{fig:GAP} illustrates this process, which we refer to as CAM-GAP.
From \cref{eq:gap_score} we observe that it encourages all pixels in the feature map to identify with the target class.
This may disadvantage small segments and increase the reliance of the classifier on the context, which can be undesirable due to a loss in mask precision.
Also, as becomes evident from \cref{eq:cam}, there are two more issues if we were to adopt CAM-GAP to provide segment cues for learning.
First, the mask value is not bounded from above, yet in segmentation we seek a normalised representation (\eg $\in (0, 1)$) that can be interpreted as a confidence by downstream applications.
Second, GAP does not encode the notion of pixel-level competition from the underlying segmentation task where each pixel can assume only one class label (\ie there is no \texttt{softmax} or a related component).
We thus argue that CAM-GAP is ill-suited for the segmentation task.

\myparagraph{Going beyond CAMs.}
To address this, we propose a novel scheme of score aggregation, see \cref{fig:GWP} for an overview, which allows for seamless integration into existing backbones, yet does not inherit the shortcomings of CAM-GAP.
Note that the following discussion is orthogonal to the loss function applied on the final classification scores, which we keep from our baseline model.

Given features $x_{:,:,:}$, we first predict classification scores $y_{:,:,:}$ of size $C \times h \times w$ for each pixel.
We then add a background channel (with a constant value) and compute a pixelwise \texttt{softmax} to obtain masks with confidence values $m_{:,:,:}$ -- this is a standard block in segmentation.
To compute a classification score, we propose \emph{normalised Global Weighted Pooling} (nGWP), defined as
\begin{equation}
\begin{aligned}
y_c^\text{nGWP} = \frac{\sum_{i,j} m_{c,i,j} y_{c,i,j}}{\epsilon + \sum_{i',j'} m_{c,i',j'}}.
\end{aligned}
\label{eq:gwp}
\end{equation}
Here, a small $\epsilon > 0$ tackles the saturation problem often observed in practice (\cf \cref{sec:supp_loss}).

As we observe from \cref{eq:gwp}, nGWP is invariant to the mask size.
This may bring advantages for small segments, but lead to inferior recall compared to the more aggressive GAP aggregation.
To encourage \emph{completeness}, we encourage increased mask size for positive classes with a penalty term:
\begin{equation}
\begin{aligned}
y_c^\text{size} = \log\Big(\lambda + \frac{1}{hw} \sum_{i,j} m_{c,i,j}\Big).
\end{aligned}
\label{eq:pen_simple}
\end{equation}
The magnitude of this penalty is controlled by a small $\lambda > 0$.
The logarithmic scale ensures that we incur a large negative value of the penalty only when the mask is near zero.
Since we decouple the influence of the class scores (captured by \cref{eq:gwp}) from that of the mask size (through \cref{eq:pen_simple}), we can apply difficulty-aware loss functions.
We generalise the penalty term in \cref{eq:pen_simple} to the focal loss \cite{LinGGHD17}, used in our final model:
\begin{equation}
\begin{aligned}
y_c^\text{size-focal} = (1-\bar{m}_c)^p \log(\lambda + \bar{m}_c), \:\; \bar{m}_c = \tfrac{1}{hw} \sum_{i,j} m_{c,i,j}.
\end{aligned}
\label{eq:pen_focal}
\end{equation}

Note that as the mask size approaches zero, $\bar{m}_c \rightarrow 0$, the penalty retains its original form, \ie \cref{eq:pen_simple}.
However, if the mask is non-zero, $p > 0$ discounts the further increase in mask size to focus on the failure cases of near-zero masks.
We compute our final classification scores as $y_c \equiv y_c^\text{nGWP} + y_c^\text{size-focal}$ and use the multi-label soft-margin loss function \cite{paszke2017automatic} used in previous work \cite{AhnK18,WeiXSJFH18} as the classification loss,
\begin{equation}
\begin{aligned}
\mathcal{L}_\text{cls}(\mathbf{y}, \mathbf{z}) = -\frac{1}{C} \sum^C_{c=1} & z_c \log{\bigg(\frac{1}{1 + e^{-y_c}} \bigg)} + \\
            & + (1 - z_c) \log \bigg( \frac{e^{-y_c}}{1 + e^{-y_c}} \bigg),
\end{aligned}
\label{eq:cls_loss}
\end{equation}
where $\mathbf{z}$ is a binary vector of ground-truth labels.
The loss encourages $y_c < 0$ for negative classes (\ie when $z_c = 0$) and $y_c > 0$ for positive classes (\ie when $z_c = 1$).

\subsection{Pixel-adaptive mask refinement}
\label{sec:mask_refinement}
While our classification loss accounts for \emph{semantic fidelity} and \emph{completeness},
the task of local mask refinement is to fulfil \emph{local consistency}: nearby regions sharing the same appearance should be assigned to the same class.

The mapping formalising this idea takes the pixel-level mask predictions $m_{:,:,:} \in (0, 1)^{(C + 1)\times h \times w}$ (note $+1$ for the background class) and considers the image $I$ to produce refined masks $m_{:,:,:}^\ast$.
Such a mapping has to be efficient, since we will use it to produce self-supervision for segmentation trained concurrently with the classification objective.
Therefore, a naive choice of GrabCut \cite{rother2004grabcut} or dense CRFs \cite{KrahenbuhlK11} would slow down the training process.

\begin{figure}[t]%
    \def\svgwidth{\linewidth}
    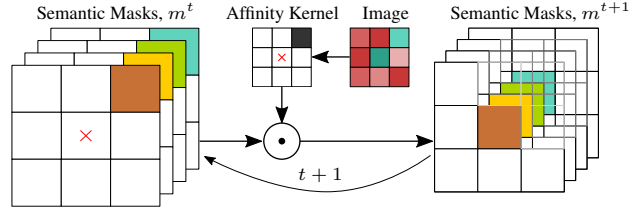
    \caption{\textbf{Concept illustration of Pixel-Adaptive Mask Refinement (PAMR).} For each pixel, we compute an affinity kernel measuring its proximity to its neighbours in RGB space. We iteratively apply this kernel to the semantic masks via an adaptive convolution to obtain refined labels.}%
    \label{fig:pixel_adaptive}%
    \vspace{-0.5em}
\end{figure}

Instead, our implementation derives from the Pixel-Adaptive Convolution (PAC) \cite{su2019pixel}.
The idea, illustrated in \cref{fig:pixel_adaptive}, is to iteratively update pixel label $m_{:,i,j}$ using a convex combination of the labels of its neighbours $\mathcal{N}(i,j)$, \ie at the $t^\text{th}$ iteration we have
\begin{equation}
\begin{aligned}
m^t_{:,i,j} = \sum_{(l,n) \in \mathcal{N}(i,j)} \alpha_{i,j,l,n} \cdot m^{t-1}_{:,l,n},
\end{aligned}
\label{eq:refine_label}
\end{equation}
where the pixel-level affinity $\alpha_{i,j,l,n}$ is a function of the image $I$.
To compute $\alpha$, we use a kernel function on the pixel intensities $I$,
\begin{equation}
\begin{aligned}
k(I_{i,j}, I_{l,n}) = -\tfrac{ \lvert I_{i,j} - I_{l,n} \rvert }{\sigma_{i,j}^2},
\end{aligned}
\end{equation}
where we define $\sigma_{i,j}$ as the standard deviation of the image intensity computed \emph{locally} for the affinity kernel.
We apply a \texttt{softmax} to obtain the final affinity distance $\alpha_{i,j,l,n}$ for each neighbour $(l,n)$ of $(i,j)$, \ie
$\alpha_{i,j,l,n} = e^{\bar{k}(I_{i,j}, I_{l,n})} / \sum_{(q,r) \in \mathcal{N}(i,j)} e^{\bar{k}(I_{i,j}, I_{q,r}))} $,
where $\bar{k}$ is the average affinity value across the RGB channels.

This local refinement, termed \emph{Pixel-Adaptive Mask Refinement} (PAMR), is implemented as a parameter-free recurrent module, which iteratively updates the labels following \cref{eq:refine_label}.
Clearly, the number of required iterations depends on the size and shape of the affinity kernel (\eg $3\times 3$ in \cref{fig:pixel_adaptive}).
In practice, we combine multiple $3 \times 3$-kernels with varying dilation rates.
We study the choice of the affinity structure in more detail in our ablation study (\cf \cref{sec:ablation}).
Note that since we do not back-propagate through PAMR, it is always in ``evaluation'' mode, hence memory-efficient.
In practice, one iteration adds less than $1\%$ of the baseline's GPU footprint, and we empirically found 10 refinement steps to provide a sufficient trade-off between the efficiency and the delivered accuracy boost from PAMR.

\begin{figure}[t]
\captionsetup[subfigure]{labelformat=empty}
\centering
\begin{subfigure}{0.19\linewidth}
\caption{Input}
\includegraphics[width=\linewidth]{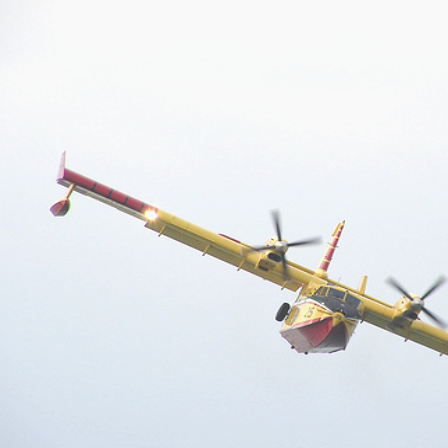}\end{subfigure}
\begin{subfigure}{0.19\linewidth}
\caption{Ground Truth}
\includegraphics[width=\linewidth]{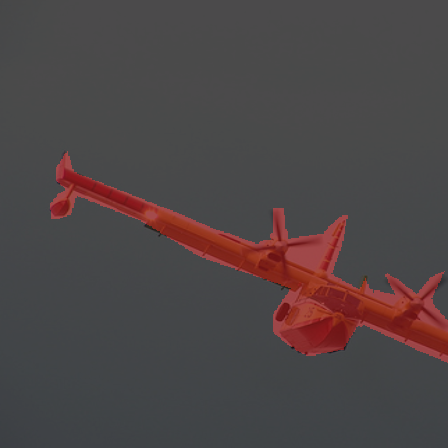}\end{subfigure}
\begin{subfigure}{0.19\linewidth}
\caption{Prediction}
\includegraphics[width=\linewidth]{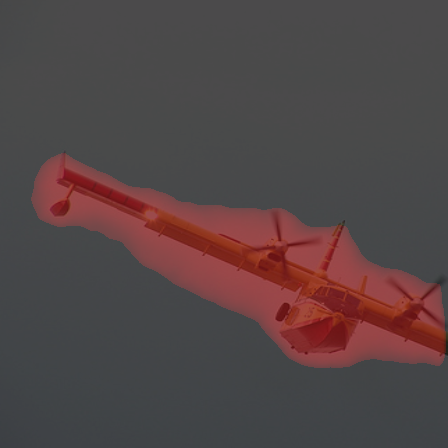}\end{subfigure}
\begin{subfigure}{0.19\linewidth}
\caption{Refinement}
\includegraphics[width=\linewidth]{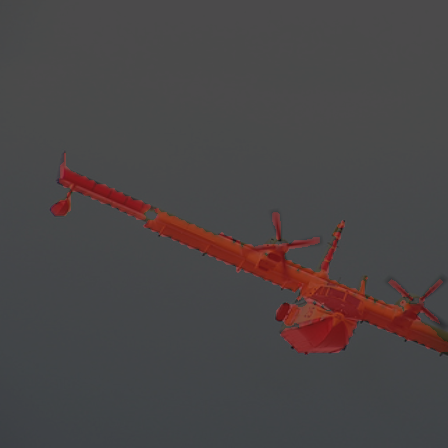}\end{subfigure}
\begin{subfigure}{0.19\linewidth}
\caption{Pseudo GT}
\includegraphics[width=\linewidth]{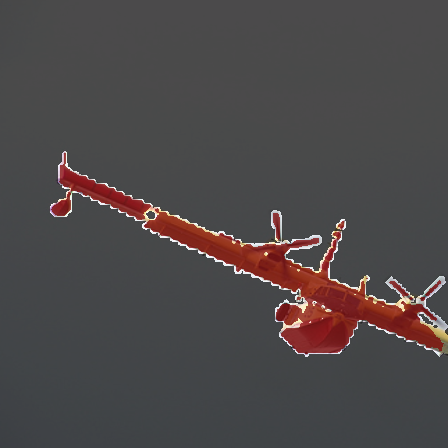}\end{subfigure}



\vspace{1mm}

\begin{subfigure}{0.19\linewidth}\includegraphics[width=\linewidth]{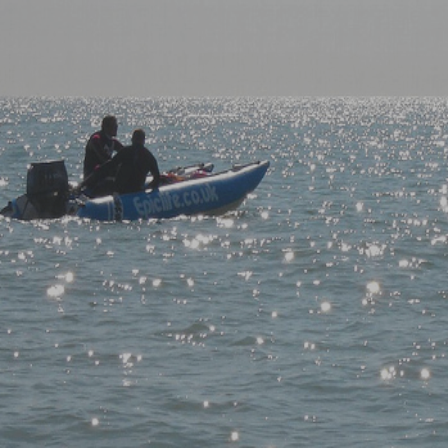}\end{subfigure}
\begin{subfigure}{0.19\linewidth}\includegraphics[width=\linewidth]{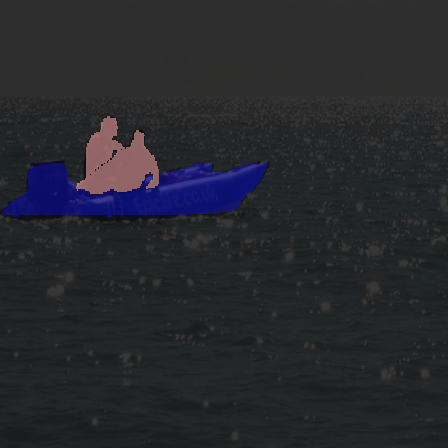}\end{subfigure}
\begin{subfigure}{0.19\linewidth}\includegraphics[width=\linewidth]{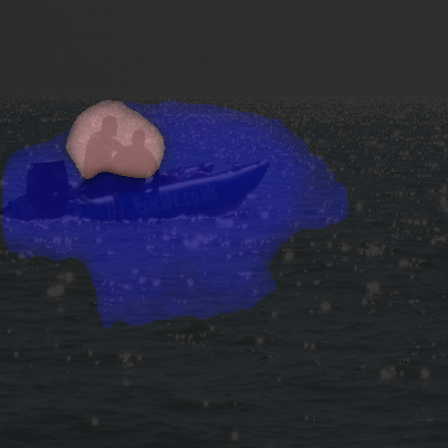}\end{subfigure}
\begin{subfigure}{0.19\linewidth}\includegraphics[width=\linewidth]{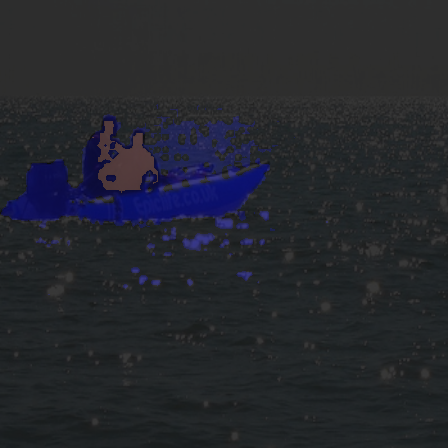}\end{subfigure}
\begin{subfigure}{0.19\linewidth}\includegraphics[width=\linewidth]{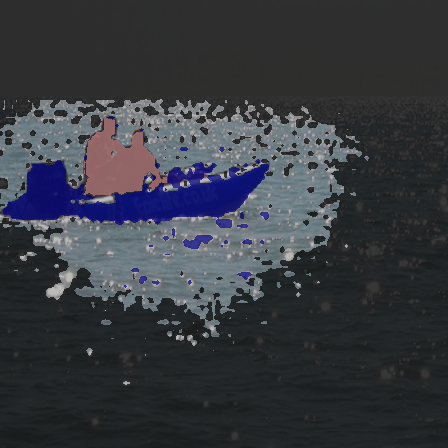}\end{subfigure}

\caption{\textbf{Intermediate results at training time.} PAMR refines model predictions \emph{(middle)} by accounting for appearance cues in the image. The revised masks \emph{($4^\text{th}$ column)} serve as pseudo ground truth, where we use only high-confidence pixels \emph{(last column)}.}
\label{fig:intermediate}
\vspace{-0.5em}
\end{figure}

\myparagraph{Self-supervised segmentation loss.} 
We generate pseudo ground-truth masks from PAMR by considering pixels with confidence $>60\%$ of the maximum value ($>70\%$ for the background class).
Conflicting pixels and pixels with low confidence are ignored by the loss function.
We fully discard images for which some of the ground-truth classes do not give rise to any confident pseudo ground-truth pixels.
Following the fully supervised case \cite{deeplabv3plus2018}, we use pixelwise cross-entropy, but balance the loss distribution across the classes, \ie the loss for each individual class is normalised \wrt the number of corresponding pixels contained in the pseudo ground truth.
The intermediate results for segmentation self-supervision at training time are illustrated in \cref{fig:intermediate}.

\subsection{Stochastic gate}
\label{sec:stochasitc_gate}
In self-supervised learning we expect the model to ``average out'' inaccuracies (manifested by their irregular nature) in the pseudo ground truth, thereby improving the predictions and thus the pseudo supervision.
However, a powerful model may just as well learn to mimic these errors.
Strong evidence from previous work indicates that the large receptive field of the deep features enables the model to learn such complex phenomena in segmentation \cite{deeplabv3plus2018,wu2019wider,zhao2017pyramid}.

\begin{figure}[t]%
    \def\svgwidth{1.01\linewidth}
    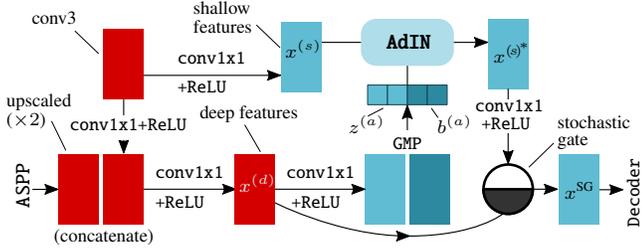
    \caption{\textbf{Concept illustration of the Stochastic Gate.} All rectangular blocks are tensors of the same size. The baseline model from DeepLabv3+ \cite{deeplabv3plus2018} is shown in red: the output from ASPP is augmented via a skip connection from \texttt{conv3} and the result, $x^{(d)}$, passes directly through the decoder. Shown in blue, our modification (GCI) infuses global cues extracted from the deep features into the shallow features via AdIN \cite{huang2017arbitrary}. The enriched shallow and the deep features are then combined using \cref{eq:bernoulli} at training and \cref{eq:gate_sum} at inference time.}%
    \label{fig:gate}%
    \vspace{-0.5em}
\end{figure}

To counter the compounding effect of the errors in self-supervision, we propose a type of regularisation, referred to as \emph{Stochastic Gate} (SG).
The underlying idea, shown in \cref{fig:gate}, is to encourage information sharing between the deep features (with a large receptive field) and ``shallow'' representation of the preceding layers (with a lower receptive field), by stochastically exchanging these representations, yet maintaining the deep features as the expected value.
Formally, let $x^{(d)}$ and $x^{(s)}$ represent the activation in the deep and shallow feature map, respectively (omitting the tensor subscripts for brevity).
Applying SG for each pixel at training time is reminiscent of Dropout \cite{SrivastavaHKSS14}:
\begin{gather}
\begin{aligned}
x^\text{SG} = (1 - r) \underbrace{\delta [ x^{(d)} - \psi x^{(s)}]}_{x^{(\ast)}} + r x^{(s)}, \:\text{with}\: r \sim \text{Bern}(\psi),
\end{aligned}
\label{eq:bernoulli}
\raisetag{10pt}
\end{gather}
where the \emph{mixing rate} $\psi \in [0, 1]$ regulates the proportion of $x^{(\ast)}$ and $x^{(s)}$ in the output tensor.
The constant $\delta=1 / (1 - \psi)$ ensures that $\mathbb{E}[x^\text{SG}] = x^{(d)}$.
It is easy to show that $x^{(\ast)} \approx x^{(s)}$, encouraged by the stochasticity, implies $x^{(d)} \approx x^{(s)}$, \ie feature sharing between the deep and shallow representations.
At inference time, we deterministically combine the two, now complementary, streams:
\begin{equation}
\begin{aligned}
x^\text{SG} = (1 - \psi) x^{(d)} + \psi x^{(s)}.
\end{aligned}
\label{eq:gate_sum}
\end{equation}

Shallow features alone may be too limited in terms of the semantic information they contain.
To enrich their representation, yet preserve their original receptive field, we devise \emph{Global Cue Injection (GCI)} via Adaptive Instance Normalisation (AdIN) \cite{huang2017arbitrary}.
As shown in \cref{fig:gate}, we first apply a $1\!\times\!1$ convolution to the deep feature tensor to double the number of channels.
Then, we extract two vectors with global information (\ie without spatial cues) via Global Max Pooling (GMP).
Shown as the left (unshaded) and right (shaded) half of the 1D vector after GMP in \cref{fig:gate}, let $z^{(a)}$ and $b^{(a)}$ denote two parts of such a representation, which will be shared by each site in a shallow feature channel.
We compute the \emph{augmented} shallow activation $x^{(s)^\ast}$ as 
\begin{equation}
\begin{aligned}
x^{(s)^\ast} = \texttt{ReLU} \Big( z^{(a)} \Big(\tfrac{x^{(s)} - \mu(x^{(s)})}{\sigma(x^{(s)})}\Big) + b^{(a)} \Big),
\end{aligned}
\end{equation}
where $\mu(\cdot)$ and $\sigma(\cdot)$ are the mean and the standard deviation of each channel of $x^{(s)}$.
The updated activation, $x^{(s)^\ast}$, goes through a $1\!\times\!1$-convolution and replaces the original $x^{(s)}$ in \cref{eq:bernoulli} and \cref{eq:gate_sum} in the final form of SG.
Following \cite{deeplabv3plus2018}, the output from SG then passes through a 3-layer decoder.

\section{Experiments}
\label{sec:exp}

\subsection{Setup}

\paragraph{Dataset.} Pascal VOC 2012~\cite{EveringhamGWWZ10} is an established benchmark for weakly supervised semantic segmentation and contains 20 object categories.
Following the standard practice~\cite{AhnK18,KolesnikovL16,WeiXSJFH18}, we augment the original VOC training data with an additional image set provided by Hariharan \etal \cite{HariharanABMM11}.
In total, we use \num{10582} images with image-level annotation for training and \num{1449} images for validation.

\myparagraph{Implementation details.}
Our model is implemented in PyTorch \cite{paszke2017automatic}.
We use a \textit{WideResNet-38} backbone network \cite{wu2019wider} provided by \cite{AhnK18} (see \cref{sec:backbones}, for experiments with VGG16 \cite{SimonyanZ14a} and ResNet backbones \cite{HeZRS16}).
We further extend this model to \textit{DeepLabv3+} by adding Atrous Spatial Pyramid Pooling (ASPP), a skip connection (with our Stochastic Gate), and the 3-layer decoder \cite{deeplabv3plus2018}.
We train our model for \num{20} epochs with SGD using weight decay \num{5e-4} with momentum \num{0.9}, a constant learning rate of \num{0.01} for the new (randomly initialised) modules and \num{0.001} for \textit{WideResNet-38} parameters, initialised from ImageNet \cite{deng2009imagenet} pre-training.
We first train our model for 5 epochs using only the classification loss and switch on the self-supervised segmentation loss for the remaining 15 epochs.
We use inference with multi-scale inputs \cite{chen2016attention} and remove masks for classes with classifier confidence $<0.1$.

\myparagraph{Data augmentation.}
Following common practice \cite{AhnK18,WeiXSJFH18}, we use random rescaling (in the $(0.9, 1.0)$ range \wrt the original image area), horizontal flipping, colour jittering, and train our model on random crops of size $321\times321$.

\begin{table}[t]
\centering
\subcaptionbox{IoU (val,\%) \wrt focal mask penalty\label{table:ablation_focal}. We fix $\psi\!=\!0.5$ w/ GCI for SG and use PAMR kernel $[1,2,4,8,12,24]$.}{%
\footnotesize%
\setlength{\tabcolsep}{0.55em}%
\begin{tabularx}{0.47\linewidth}{@{}X@{}S[table-format=2.1]S[table-format=2.1]S[table-format=2.1]@{}}
\toprule
$\downarrow p \hfill/\hfill \lambda\rightarrow$ & {$0.1$} & {$0.01$} & {$0.001$} \\
\midrule
$0$ & 58.8 & 58.9 & 57.4 \\
$3$ & 59.5 & 59.4 & 58.1 \\
$5$ & 60.2 & 59.1 & 57.1 \\
\bottomrule
\end{tabularx}
}\hfill
\subcaptionbox{IoU (val,\%) \wrt Pixel-Adaptive Mask Refinement\label{table:ablation_pamr}. We fix $\psi\!=\!0.5$ w/ GCI for SG and set $p\!=\!3, \lambda\!=\!0.01$.}{%
\footnotesize%
\setlength{\tabcolsep}{0.55em}%
\begin{tabularx}{0.47\linewidth}{@{}cccccX@{}S[table-format=2.1]@{}}
\toprule
1 & 2 & 4 & 8 & 12 & 24 & {IoU} \\
\midrule
\cmark  & \cmark  & \cmark  & \cmark  & \cmark & \cmark  & \bfseries 59.4 \\
\midrule
\multicolumn{6}{@{}l}{\footnotesize{\textit{(no refinement)}}} & 31.8 \\
\cmark  & \cmark  & \cmark  & \cmark  &  &  & 50.6 \\
\cmark  & \cmark  & \cmark  & \cmark  & \cmark &    & 55.7 \\
\cmark  & 3  & 6  &  9  & \cmark &  16  & 57.9 \\
\cmark  & \cmark  & \cmark  & \cmark  & \cmark & 16 & 58.2 \\
\bottomrule
\end{tabularx}
}\\[0.5em]
\subcaptionbox{IoU (val,\%) \wrt the Stochastic Gate\label{table:ablation_gate}. We use $p\!=\!3, \lambda\!=\!0.01$ for the focal penalty and use PAMR kernel $[1,2,4,8,12,24]$.}{%
\footnotesize
\begin{tabularx}{\linewidth}{@{}XS[table-format=2.1]@{\hspace{2em}}S[table-format=2.1]@{}}
\toprule
Config & {IoU} & {IoU \scriptsize{(+ CRF)}} \\
\midrule
$\psi = 0.5$ & 59.4 & 62.2 \\
$\psi = 0.5$ w/o GCI & \bfseries 59.8 & 60.9 \\
\midrule
$\psi = 0.3$ & \bfseries 59.7 & \bfseries 62.7 \\
$\psi = 0.3$ w/o GCI & 57.7 & 60.3 \\
\midrule
w/o SG & 55.6 & 57.5 \\
Deterministic Gate & 57.5 & 57.7 \\
\bottomrule
\end{tabularx}
}
\caption{\textbf{Ablation study on Pascal VOC.} We study the role of \subref{table:ablation_focal} the focal mask penalty, \subref{table:ablation_pamr} the Pixel-Adaptive Mask Refinement, and \subref{table:ablation_gate} the Stochastic Gate.}
\label{table:ablation}
\vspace{-0.5em}
\end{table}

\subsection{Ablation study}
\label{sec:ablation}

\paragraph{Focal mask penalty.}
Following the intuition from \cite{LinGGHD17}, the focal mask penalty emphasises training on the current failure cases, \ie small (large) masks for the classes present (absent) in the image.
Recall from \cref{eq:pen_focal} that $\lambda$ controls the penalty magnitude, while $p$ is the discounting rate for better-off image samples.
We aim to verify if the ``focal'' aspect of the mask penalty provides advantages over the baseline penalty (\ie $p = 0$).
\cref{table:ablation_focal} summarises the results.

First, we find that the focal version of the mask penalty improves the segmentation quality of the baseline.
This improvement, maximised with $p=5$ and $\lambda=0.1$, is tangible, yet comes at a negligible computational cost.
Second, we observe that increasing $\lambda$ tends to increase the segmentation accuracy.
While changing $\lambda$ from $0.01$ to $0.001$ leads to higher recall on average, it has a detrimental effect on precision.
Lastly, we also find that moderate positive values of $p$ in conjunction with CRF refinement lead to more sizeable gains in mask quality:
with $p=3, \lambda=0.01$ we achieve $62.2\%$ IoU, whereas the highest IoU with $p=0$ is only $60.5\%$ (reached with $\lambda=0.01$).
However, higher values of $p$ do not benefit from CRF processing (\eg $59.8\%$ with $p=5, \lambda=0.1$).
Hence, $p=3$ strikes the best balance between the model accuracy with and without using a CRF.
Note that removing the mask penalty, $y_c^\text{size-focal}$, leads to an expected drop in recall, reaching only $56.6\%$ IoU.

\begin{table}
\footnotesize
\begin{tabularx}{\linewidth}{@{}XS[table-format=2.1]@{\hspace{2em}}S[table-format=2.1]@{}}
\toprule
Method & {IoU (train,\%)} & {IoU (val,\%)} \\
\midrule
CAM \cite{AhnK18} \scriptsize{(Our Baseline)} & 48.0 & 46.8 \\
CAM + RW \cite{AhnK18} & 58.1 & 57.0 \\
CAM + RW + CRF \cite{AhnK18} & 59.7 & {--} \\
CAM + IRN + CRF \cite{AhnCK19} & 66.5 & {--} \\
\toprule
Ours & 64.7 & 63.4 \\
Ours + CRF & \bfseries 66.9 & \bfseries 65.3 \\
\bottomrule
\end{tabularx}
\caption{\textbf{Segmentation quality on Pascal VOC training and validation sets.} Here, we use ground-truth image-level labels to remove masks of any   false positive classes predicted by our model.}
\label{table:seg_training}
\vspace{-0.5em}
\end{table}

\myparagraph{Pixel-Adaptive Mask Refinement (PAMR).}
Recall from \cref{sec:mask_refinement} that PAMR aims to improve the quality of the original coarse masks \wrt \emph{local consistency} to provide self-supervision for segmentation.
Here, we verify \textit{(i)} the importance of PAMR by training our model without the refinement; and \textit{(ii)} the choice of the kernel structure, \ie the composition of dilation rates of the $3\times3$-kernels in PAMR.

The results in \cref{table:ablation_pamr} show that PAMR is a crucial component in our self-supervised model, as the segmentation accuracy drops markedly from $59.4 \%$ to $31.8 \%$ without it.
We find further that the size of the kernel also affects the accuracy.
This is expected, since small receptive fields (dilations \texttt{1-2-4-8} in \cref{table:ablation_pamr}) are insufficient to revise the boundaries of the coarse masks that typically exhibit large deviations from the object boundaries.
The results with larger receptive fields of the affinity kernel further support this intuition: increasing the dilation of the largest $3\!\times\!3$-kernel to $24$ attains the best mask quality compared to the smaller affinity kernels.
Furthermore, we observe that varying the kernel \emph{shape} does not have such a drastic effect;
the change from \texttt{1-3-6-9-12-16} to \texttt{1-2-4-8-12-16} only leads to small accuracy changes.
This is desirable in practice as sensitivity to these minor details would imply that our architecture overfits to particularities in the data \cite{torralba2011unbiased}.

\myparagraph{Stochastic Gate (SG).}
The intention of the SG, introduced in \cref{sec:stochasitc_gate}, is to counter overfitting to the errors contained in the pseudo supervision.
Here, there are four baselines we aim to verify: \emph{(i)} disabling SG; \emph{(ii)} combining $x^{(d)}$ and $x^{(s)}$ deterministically (\ie $r\equiv\psi$ in \cref{eq:bernoulli}); \emph{(iii)} the role of the Global Cue Injection (GCI); and \emph{(iv)} the effect of the mixing rate $\psi$.
These results are summarised in \cref{table:ablation_gate}.
Evidently, SG is crucial, since disabling it substantially weakens the mask accuracy (from 59.8\% to 55.6\% IoU).
The \emph{stochastic} nature of SG is also important: simply summing up $x^{(d)}$ and $x^{(s)}$ (we used $r\equiv\psi=0.5$) yields inferior mask IoU (57.5\% \vs 59.8\%).
In our model comparison with both $\psi=0.5$ and $\psi=0.3$, we find that the model with GCI tends to provide superior results.
However, the model without GCI can be as competitive given a particular choice of $\psi$ (\eg, $0.5$).
In this case, the model with GCI usually has higher recall, while the model without it has higher precision.
Since CRFs tend to increase the precision provided sufficient mask support, the model with GCI should therefore profit more from this refinement.
We confirmed this and observed a more sizeable improvement of the model with GCI (59.4 \vs 62.2\% IoU).
Additionally, we found GCI to deliver more stable results for different $\psi$, which can alleviate parameter fine-tuning in practice.

\begin{table}
\footnotesize
\begin{tabularx}{\linewidth}{@{}X@{\hspace{0.5em}}l@{\hspace{0.25em}}c@{\hspace{0.25em}}c@{\hspace{1em}}S[table-format=2.1]@{\hspace{1em}}S[table-format=2.1]@{}}
\toprule
Method & Backbone & Superv. & Dep. & {val} & {test} \\
\midrule
\multicolumn{6}{@{}l}{\scriptsize \textit{Fully supervised}} \\
\midrule
WideResNet38~\cite{wu2019wider} & & $\mathcal{F}$ &  & 80.8 & 82.5 \\
DeepLabv3+~\cite{deeplabv3plus2018} & Xception-65 \cite{chollet17} & $\mathcal{F}$ &  & {--} & 87.8 \\
\midrule
\multicolumn{6}{@{}l}{\scriptsize \textit{Multi stage + Saliency}} \\
\midrule
STC \cite{WeiLCSCFZY17} & DeepLab \cite{chen2017deeplab} & $\mathcal{S}, \mathcal{D}$ & \cite{JiangWYWZL13} & 49.8 & 51.2 \\
SEC \cite{KolesnikovL16} & VGG-16 & $\mathcal{S}, \mathcal{D}$ & \cite{SimonyanVZ13} & 50.7 & 51.7 \\
Saliency \cite{OhBKAFS17} & VGG-16 & $\mathcal{S}, \mathcal{D}$ & & 55.7 & 56.7 \\
DCSP \cite{ChaudhryDT17} & ResNet-101 & $\mathcal{S}$ & \cite{LiuH16} & 60.8 & 61.9 \\
RDC \cite{WeiXSJFH18} & VGG-16 & $\mathcal{S}$ & \cite{xiao2017self} & 60.4 & 60.8 \\
DSRG \cite{HuangWWLW18} & ResNet-101 & $\mathcal{S}$ & \cite{WangJYCHZ17} & 61.4 & 63.2 \\
FickleNet~\cite{LeeKLLY19} & ResNet-101 & $\mathcal{S}$ & \cite{HuangWWLW18} & 64.9 & 65.3 \\
\multirow{2}{*}{Frame-to-Frame~\cite{Lee_2019_ICCV}} & VGG-16 & \multirow{2}{*}{$\mathcal{S}, \mathcal{D}$} & \multirow{2}{*}{\cite{sun2018pwc,HouCHBTT17,LeeKLLY19}} & 63.9 & 65.0 \\
 & ResNet-101 & & & 66.5 & 67.4 \\
\midrule
\multicolumn{6}{@{}l}{\scriptsize \textit{Single stage + Saliency}} \\
\midrule
JointSaliency~\cite{Zeng_2019_ICCV} & \scriptsize{DenseNet-169} \cite{HuangLMW17} & $\mathcal{S}, \mathcal{D}$ & & 63.3 & 64.3 \\
\midrule
\multicolumn{6}{@{}l}{\scriptsize \textit{Multi stage}} \\
\midrule
AffinityNet~\cite{AhnK18} & WideResNet-38 & $\mathcal{I}$ & & 61.7 & 63.7 \\
IRN~\cite{AhnCK19} & ResNet-50 & $\mathcal{I}$ & & 63.5 & 64.8 \\
SSDD~\cite{Shimoda_2019_ICCV} & WideResNet-38 & $\mathcal{I}$ & \cite{AhnK18} & 64.9 & 65.5 \\
\midrule
\multicolumn{6}{@{}l}{\scriptsize \textit{Single stage}} \\
\midrule
TransferNet~\cite{HongOLH16} & VGG-16 & $\mathcal{D}$ & & 52.1 & 51.2 \\
WebCrawl~\cite{HongYKLH17} & VGG-16 & $\mathcal{D}$ & & 58.1 & 58.7 \\
\midrule
EM~\cite{PapandreouCMY15} & VGG-16 & $\mathcal{I}$ &  & 38.2 & 39.6 \\
MIL-LSE~\cite{PinheiroC15} & Overfeat \cite{SermanetEZMFL13} & $\mathcal{I}$ & & 42.0 & 40.6 \\
CRF-RNN \cite{RoyT17} & VGG-16 & $\mathcal{I}$ &  & 52.8 & 53.7 \\
\midrule 
Ours & \multirow{2}{*}{WideResNet-38} & $\mathcal{I}$ & & 59.7 & 60.5 \\
Ours + CRF &  & $\mathcal{I}$ & &  62.7 & 64.3 \\
\bottomrule
\end{tabularx}
\caption{\textbf{Mean IoU (\%) on Pascal VOC validation and test.}
For each method we indicate additional cues used for training beyond image-level labels $\mathcal{I}$, such as saliency detection ($\mathcal{S}$), additional data $(\mathcal{D})$, as well as their dependence on other methods (``Dep.'').
}
\label{table:main_result}
\vspace{-0.5em}
\end{table}

\subsection{Comparison to the state of the art}
\label{sec:main_result}
\paragraph{Setup.} Here, our model uses SG with GCI, $\psi=0.3$, the focal penalty with $p=3$ and $\lambda=0.01$, and PAMR with $10$ iterations and a \texttt{1-2-4-8-12-24} affinity kernel.

\begin{figure*}[t]
    \centering
    \begin{subfigure}[t]{.375\linewidth}
          \centering\large
            \includegraphics[width=0.98\linewidth]{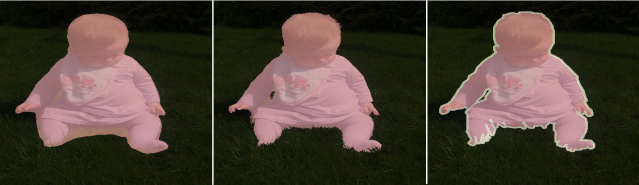}\\
      \includegraphics[width=0.98\linewidth]{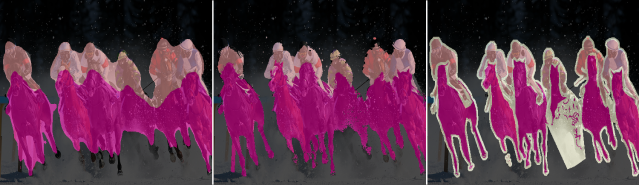}\\
      \includegraphics[width=0.98\linewidth]{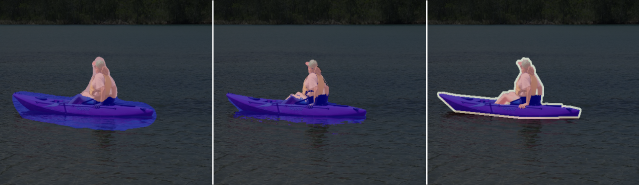}\\
      \includegraphics[width=0.98\linewidth]{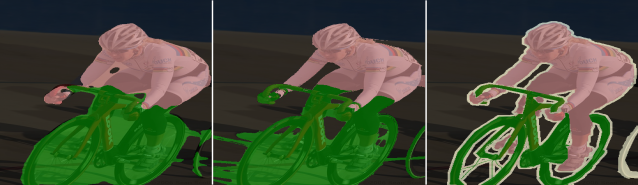}
          \caption{Train}\label{fig:q_train}
    \end{subfigure}%
    \begin{subfigure}[t]{.375\linewidth}
          \centering\large
            \includegraphics[width=0.976\linewidth]{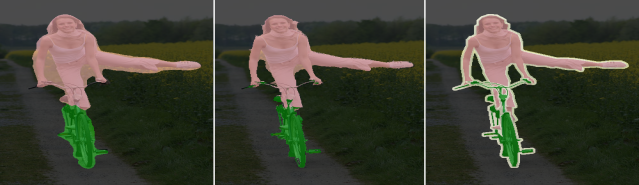}\\
      \includegraphics[width=0.976\linewidth]{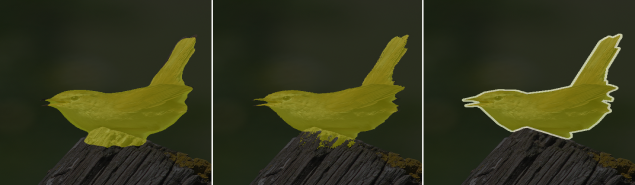}\\
      \includegraphics[width=0.976\linewidth]{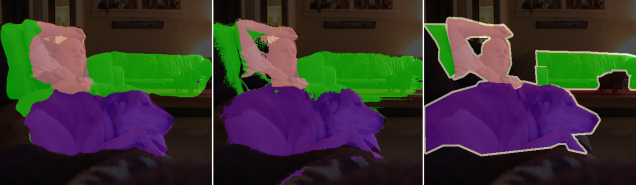}\\
      \includegraphics[width=0.976\linewidth]{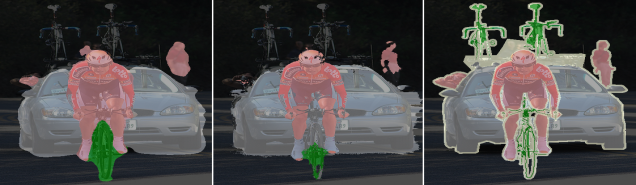}
          \caption{Val}\label{fig:q_val}
    \end{subfigure}%
    \begin{subfigure}[t]{.25\linewidth}
          \centering\large
            \includegraphics[width=0.976\linewidth]{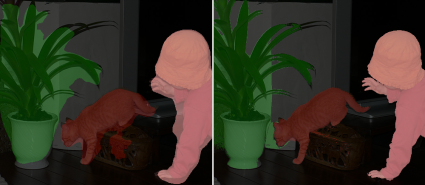}\\
      \includegraphics[width=0.976\linewidth]{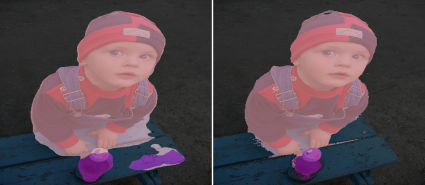}\\
      \includegraphics[width=0.976\linewidth]{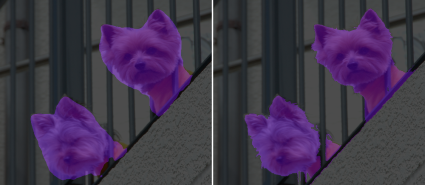}\\
      \includegraphics[width=0.976\linewidth]{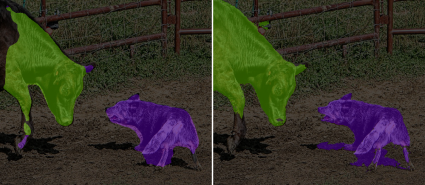}
          \caption{Test}\label{fig:q_test}
    \end{subfigure}

    \caption{\textbf{Qualitative results on PASCAL VOC.} We show example segmentations from our method (\textit{left}), the result of CRF post-processing (\textit{middle}), and the ground truth (\textit{right}). Our method produces masks of high quality under a variety of challenging conditions.}
    \label{fig:qualitative}%
    \vspace{-0.5em}
\end{figure*}

\myparagraph{Mask quality.}
Recall that the majority of recent work, \eg \cite{HuangWWLW18,LeeKLLY19,Lee_2019_ICCV,WeiXSJFH18}, additionally trains a separate fully supervised segmentation network from the pseudo ground truth.
To evaluate the quality of such pseudo supervision generated by our model, we use image-level ground-truth labels to remove any masks for classes that are not present in the image (for this experiment only).
The results in \cref{table:seg_training} show that using our single-stage mask output as pseudo segmentation labels improves the baseline IoU of CAMs by a staggering $18.9 \%$ IoU ($18.5 \%$ on validation), and even outperforms recent multi-stage methods \cite{AhnK18} by $7.2\%$ IoU and \cite{AhnCK19} by $0.4\%$ IoU.
This is remarkable, since we neither train additional models nor recourse to saliency detection.

\myparagraph{Segmentation accuracy.}
\cref{table:main_result} provides a comparative overview \wrt the state of the art.
Since image-level labels are generally not available at test time, we do \emph{not} perform any mask pruning here (unlike \cref{table:seg_training}).
In the setting of image-level supervision, IRN \cite{AhnCK19} and SSDD \cite{Shimoda_2019_ICCV} are the only methods with higher IoU than ours.
Both methods are multi-stage; they are trained in at least \emph{three} stages.
IRNet \cite{AhnCK19} trains an additional segmentation network on pseudo labels to eventually outperform our method by only $0.5\%$ IoU.
Recall that SSDD is essentially a post-processing approach: it refines the masks from AffinityNet \cite{AhnK18} (with $63.7\%$ IoU) using an additional network and further employs a cascade of \emph{two} networks to revise the masks.
This strategy improves over our results by only $1.2\%$ IoU, yet at the cost of a considerable increase in model complexity.

Our single-stage method is also competitive with JointSaliency \cite{Zeng_2019_ICCV}, which uses a more powerful backbone \cite{HuangLMW17} and saliency supervision.
The recent Frame-to-Frame system \cite{Lee_2019_ICCV} is also supervised with saliency and is trained on additional 15K images mined from videos, which requires state-of-the-art optical flow \cite{sun2018pwc}.
By contrast, our approach is substantially simpler since we train \emph{one} network in a \emph{single} shot.
Nevertheless, we surpass a number of multi-stage methods that use additional data and saliency supervision \cite{ChaudhryDT17,HuangWWLW18,KolesnikovL16,OhBKAFS17,WeiLCSCFZY17}.
We significantly improve over previous single-stage methods \cite{PapandreouCMY15,PinheiroC15,RoyT17},
as well as outperform the single-stage WebCrawl \cite{HongYKLH17}, which relies on additional training data and needs multiple forward passes through its class-agnostic decoder.
Our model needs neither and infers masks for all classes in one pass.

Note that training a standalone segmentation network on our pseudo labels is a trivial extension, which we omit here in view of our practical goals.
However, we still provide these results in the supplemental material (\cref{sec:pseudo_labels}), in fact, achieving state of the art in a \emph{multi-stage} setup as well.

\myparagraph{Qualitative analysis.}
From the qualitative results in \cref{fig:qualitative}, we observe that our method produces segmentation masks that align well with object boundaries.
Our model exhibits good generalisation to challenging scenes with varying object scales and semantic content.
Common failure modes of our segmentation network are akin to those of fully supervised methods: segmenting fine-grained details (\eg, bicycle wheels), mislabelling under conditions of occlusions (\eg, leg of the cyclist \vs bicycle), and misleading appearance cues (\eg, low contrast, similar texture).

\section{Conclusion}

In this work, we proposed a practical approach to weakly supervised semantic segmentation, which comprises a single segmentation network trained in one round.
To ensure local consistency, semantic fidelity, and completeness of the segmentation masks, we introduced a new class aggregation function, a local mask refinement module, and a stochastic gate.
Our approach is astonishingly effective despite its simplicity.
Specifically, it yields segmentation accuracy on par with the state of the art and outperforms a range of recent multi-stage methods relying on additional training data and saliency supervision.
We expect that our model can also profit from auxiliary supervision and suit the needs of downstream tasks without considerable deployment effort.

{\small
\bibliographystyle{ieee_fullname}
\bibliography{egbib}
}

\clearpage
\pagenumbering{roman}
\appendix

\title{Single-Stage Semantic Segmentation from Image Labels\\\large -- Supplemental Material --}
\author{Nikita Araslanov \hspace{1cm} Stefan Roth\\
Department of Computer Science, TU Darmstadt}

\maketitle


\section{On Training Stages}
\label{sec:stages}

\cref{table:related_work} provides a concise overview of previous work on semantic segmentation from image-level labels.
As an important practical consideration, we highlight the number of stages used by each method.
In this work, we refer to one stage as learning an independent set of model parameters with intermediate results saved \emph{offline} as input to the next stage.
For example, the approach by Ahn~\etal \cite{AhnK18} comprises three stages: \emph{(1)} extracting CAMs as seed masks; \emph{(2)} learning a pixel affinity network to refine these masks; and \emph{(3)} training a segmentation network on the pseudo labels generated by affinity propagation.
Note that three methods \cite{JingCT20,WeiFLCZY17,WeiLCSCFZY17} also use multiple training cycles (given in parentheses) of the same model, which essentially acts as a multiplier \wrt the total number of stages.
Finally, we note if a method relies on saliency detection, extra data, or previous frameworks.
We observe that predominantly early single-stage methods stand in contrast to the more complex recent multi-stage pipelines.
Our approach is single stage and relies neither on previous frameworks, nor on saliency supervision.\footnote{The background cues provided by saliency detection methods give a substantial advantage, since $63\%$ of 10K train images (VOC+SBD) have only one class.}

\begin{table}
\footnotesize
\begin{tabularx}{\linewidth}{@{}Xllc@{}} 
\toprule
Method & Extras & \# of Stages & PSR \\
\midrule
MIL-FCN \begin{tiny}\textit{ICLR '15}\end{tiny} \cite{PathakSLD14} & -- & 1 & \xmark \\
MIL-LSE \begin{tiny}\textit{CVPR '15}\end{tiny} \cite{PinheiroC15} & -- &  1 & \xmark  \\
EM \begin{tiny}\textit{ICCV '15}\end{tiny} \cite{PapandreouCMY15} & -- &  1 & \xmark \\
TransferNet \begin{tiny}\textit{CVPR '16}\end{tiny} \cite{HongOLH16} & -- & 1 & \xmark  \\
SEC \begin{tiny}\textit{ECCV '16}\end{tiny} \cite{KolesnikovL16} & $\mathcal{S}$ \cite{SimonyanVZ13} & 2 & \xmark  \\
DCSP \begin{tiny}\textit{ECCV '17}\end{tiny} \cite{ChaudhryDT17} & $\mathcal{S}$ \cite{LiuH16} & 1 & \xmark  \\
AdvErasing \begin{tiny}\textit{CVPR '17}\end{tiny} \cite{WeiFLCZY17} & $\mathcal{S}$ \cite{WangJYCHZ17} & 2 $(\times 3)$ & \cmark \\
WebCrawl \begin{tiny}\textit{CVPR '17}\end{tiny} \cite{HongYKLH17} & $\mathcal{D}$ & 1 & \xmark \\
CRF-RNN \begin{tiny}\textit{CVPR '17}\end{tiny} \cite{RoyT17} & -- & 1 & \xmark \\
STC \begin{tiny}\textit{TPAMI '17}\end{tiny} \cite{WeiLCSCFZY17} & $\mathcal{D}$, $\mathcal{S}$ \cite{JiangWYWZL13} & 3 $(\times3)$ & \cmark \\
MCOF \begin{tiny}\textit{CVPR '18}\end{tiny} \cite{wang2018weakly} & $\mathcal{S}$ \cite{wang2018weakly} & 2 & \xmark \\
RDC \begin{tiny}\textit{CVPR '18}\end{tiny} \cite{WeiXSJFH18} & $\mathcal{S}$ \cite{xiao2017self} & 3 & \cmark \\
DSRG \begin{tiny}\textit{CVPR '18}\end{tiny} \cite{HuangWWLW18} & $\mathcal{S}$ \cite{WangJYCHZ17} & 2 & \cmark \\
Guided-Att \begin{tiny}\textit{CVPR '18}\end{tiny} \cite{li2018tell} & SEC \cite{KolesnikovL16} & 1+2 & \xmark \\
SalientInstances \begin{tiny}\textit{ECCV '18}\end{tiny} \cite{fan2018associating} & $\mathcal{S}$ \cite{fan2019s4net} & 2 & \cmark \\
Affinity \begin{tiny}\textit{CVPR '18}\end{tiny} \cite{AhnK18} & -- & 3 & \cmark \\
SeeNet \begin{tiny}\textit{NIPS '18}\end{tiny} \cite{HouJWC18} & $\mathcal{S}$ \cite{HouCHBTT19} & 2 & \cmark \\
FickleNet \begin{tiny}\textit{CVPR '19}\end{tiny} \cite{LeeKLLY19}  & $\mathcal{S}$, DSRG \cite{HuangWWLW18} & 1+3 & \cmark \\
JointSaliency \begin{tiny}\textit{ICCV '19}\end{tiny} \cite{Zeng_2019_ICCV} & $\mathcal{S}$ & 1 & \xmark \\
Frame-to-Frame \begin{tiny}\textit{ICCV '19}\end{tiny} \cite{Lee_2019_ICCV} & $\mathcal{D}$, $\mathcal{S}$ \cite{HouCHBTT17} & 2 & \cmark \\
SSDD \begin{tiny}\textit{ICCV '19}\end{tiny} \cite{Shimoda_2019_ICCV} & Affinity \cite{AhnK18} & 2+3 & \cmark \\
IRN \begin{tiny}\textit{CVPR '19}\end{tiny} \cite{AhnCK19} & -- & 3 & \cmark \\
Coarse-to-Fine \begin{tiny}\textit{TIP '20}\end{tiny} \cite{JingCT20} & GrabCut \cite{rother2004grabcut} & 2 $(\times 5)$ & \cmark \\
\midrule
Ours & -- & 1 & \xmark \\
\bottomrule
\end{tabularx}
\caption{\textbf{Summary of related work.} We analyse the related methods \wrt external input (``Extras''), such as saliency detection ($\mathcal{S}$), additional data $(\mathcal{D})$, or their reliance on previous work. We count the number of training stages in the method and note in parentheses if the method uses multiple training cycles of the same models. We also mark methods that additionally train a standalone segmentation network in a (pseudo) fully supervised regime to refine the masks (PSR).}
\label{table:related_work}
\end{table}

\section{Loss Functions}
In this section, we take a detailed look at the employed loss functions.
Since we compute the classification scores differently from previous work, we also provide additional analysis to justify the form of this novel formulation.

\label{sec:supp_loss}
\subsection{Classification loss}
\begin{figure}[t]
\input{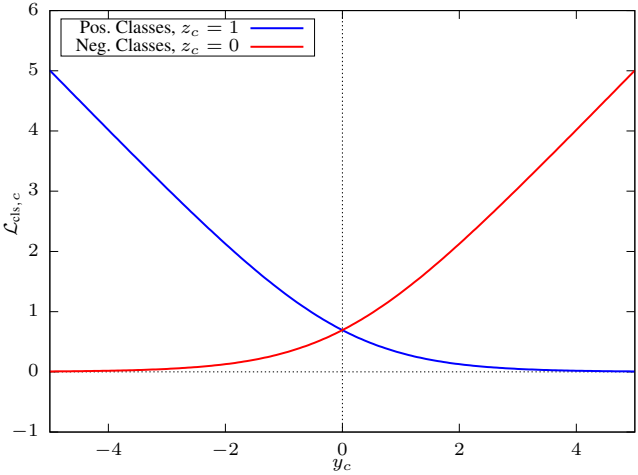}
\caption{\textbf{Soft-margin classification loss.} The loss encourages $y_c>0$ for positive classes and $y_c<0$ for negative classes. The function exhibits saturation regions: as $y_c \rightarrow \infty$ for positive classes, the associated loss (shown in blue) approaches $0$. Conversely, the loss for negative classes, shown in red, approaches $0$, as $y_c \rightarrow -\infty$.}
\label{fig:supp_cls_loss}
\end{figure}

We use the multi-label soft-margin loss function used in previous work \cite{AhnK18,WeiXSJFH18} as the classification loss, $\mathcal{L}_\text{cls}$.
Given model predictions $\mathbf{y} \in \mathbb{R}^C$ (\cf \cref{eq:gwp} and \cref{eq:pen_focal}; see also below) and a binary vector of ground-truth labels \mbox{$\mathbf{z}\in\{0,1\}^C$}, we compute the multi-label soft-margin loss \cite{paszke2017automatic} as
\begin{multline}
\mathcal{L}_\text{cls}(\mathbf{y}, \mathbf{z}) = -\frac{1}{C} \sum^C_{c=1} z_c \log{\bigg(\frac{1}{1 + e^{-y_c}} \bigg)} + \\
		 + (1 - z_c) \log \bigg( \frac{e^{-y_c}}{1 + e^{-y_c}} \bigg).
\label{eq:supp_cls_loss}
\end{multline}
As illustrated in \cref{fig:supp_cls_loss}, the loss function encourages \mbox{$y_c < 0$} for negative classes (\ie when $z_c = 0$) and $y_c > 0$ for positive classes (\ie when $z_c = 1$).
This observation will be useful for our discussion below.

Recall from \cref{eq:gwp} and \cref{eq:pen_focal} that we define our classification score $y_c$ as
\begin{equation}
y_c = y_c^\text{nGWP} + y_c^\text{size-focal}.
\label{eq:supp_yc}
\end{equation}
For convenience, we re-iterate the definition of the two terms, the normalised Global Weighted Pooling
\begin{align}
y_c^\text{nGWP} &= \frac{\sum_{i,j} m_{c,i,j} y_{c,i,j}}{\epsilon + \sum_{i',j'} m_{c,i',j'}}
\label{eq:supp_gwp}\\
\intertext{and the focal penalty}
y_c^\text{size-focal} &= (1 - \bar{m}_c)^p \log(\lambda + \bar{m}_c), \label{eq:supp_fp}\\ 
\text{with}\qquad \bar{m}_c &= \frac{1}{hw} \sum_{i,j} m_{c,i,j}.
\end{align}
Note that $y_{c,i,j}$ in \cref{eq:supp_gwp} refers to pixel site $(i, j)$ on the score map of class $c$, whereas $y_c$ in \cref{eq:supp_yc} is the \emph{aggregated} score for class $c$.
We compute the class confidence $m_{:,i,j}$ for site $(i, j)$ using a \texttt{softmax} on $y_{:,i,j}$, where we include an additional \emph{background} channel $y_{0,i,j}$.
We fix $y_{0,:,:}\equiv 1$ throughout our experiments.

The use of hyperparameter $\epsilon > 0$ in the definition of \cref{eq:supp_gwp} may look rather arbitrary and redundant at first.
In the analysis below, we show that in fact it serves multiple purposes:
\begin{enumerate}
\item \textbf{Numerical stability}. First, using $\epsilon > 0$ prevents division by zero for saturated scores, \ie where $\sum_{i',j'} m_{c,i',j'} = 0$ for some $c$ in the denominator of \cref{eq:supp_gwp}, which can happen (approximately) in the course of training for negative classes.

Secondly, $\epsilon > 0$ resolves discontinuity issues.
Observe that
\begin{equation}
\lim_{m_{c,:,:} \rightarrow 0} \:\frac{\sum_{i,j} m_{c,i,j} y_{c,i,j}}{\epsilon + \sum_{i',j'} m_{c,i',j'}} = 0,
\label{eq:supp_limit_zero}
\end{equation}
\ie $y_c^\text{nGWP} \approx 0$ for negative classes and positive $\epsilon$.

However, with $\epsilon=0$ the nGWP term in \cref{eq:supp_gwp} is not continuous at 0, which in practice may result in unstable training when $\sum_{i,j} m_{c,i,j}  \approx 0$ for some $c$.
One exception, unlikely to occur in practice, is when $y_{c,i,j} = y_{c,k,l} = d$ for all $(i,j)$ and $(k,l)$.
Then,
\begin{multline}
\lim_{m_{c,:,:} \rightarrow 0} \frac{\sum_{i,j} m_{c,i,j} y_{c,i,j}}{\sum_{i',j'} m_{c,i',j'}} = \\
= \lim_{m_{c,:,:} \rightarrow 0} \frac{d \sum_{i,j} m_{c,i,j}}{\sum_{i',j'} m_{c,i',j'}} = d.
\end{multline}
In the case of more practical relevance, \ie when $y_{c,i,j} \neq y_{c,k,l}$ for some $(i,j) \neq (k,l)$, the limit does not exist, as the following lemma shows.
With some abuse of notation, we here write $m_i$ and $y_i$ to refer to the confidence and the score values of class $c$ at pixel site $i$, respectively.
\paragraph{Lemma 1.} Let $\epsilon = 0$ in \cref{eq:supp_gwp} and suppose there exist $k$ and $l$ such that $y_k \neq y_l$.
Then, the corresponding limit
\begin{equation}
\begin{aligned}
\lim_{m_i \rightarrow 0, \forall i} \frac{\sum_i m_i y_i}{\sum_{i'} m_{i'}}
\end{aligned}
\label{eq:supp_limit_lemma}
\end{equation}
does not exist.

\begin{proof}
Let $m_k(t) = t$ and $m_i(t) = 0$ for $i \neq k$.
Then,
$$
\lim_{m_i \rightarrow 0, \forall i} \frac{\sum_i m_i y_i}{\sum_{i'} m_{i'}} = \lim_{t \rightarrow 0} \frac{t y_k}{t} = y_k.
$$
On the other hand, if we let $m_l(t) = t$ and $m_i(t) = 0$ for $i \neq l$, we obtain
$$
\lim_{m_i \rightarrow 0, \forall i} \frac{\sum_i m_i y_i}{\sum_{i'} m_{i'}} = \lim_{t \rightarrow 0} \frac{t y_l}{t} = y_l.
$$
Since $y_k \neq y_l$ by our assumption, we have now found two paths of the multivariable limit in \cref{eq:supp_limit_lemma}, which evaluate to different values.
Therefore, the limit in \cref{eq:supp_limit_lemma} is not unique, \ie does not exist (\citesupp{stewart12}, Ch.~14.2).
\end{proof}

\item \textbf{Emphasis on the focal penalty}. 
For negative classes, it is not sensible to give meaningful relative weightings of the pixels for nGWP in \cref{eq:supp_gwp}; we seek such a relative weighting of different pixels only for positive classes.
At training time we would thus like to minimise the class scores for negative classes \emph{uniformly} for all pixel sites.
Empirically, we observed that with $\epsilon > 0$ the focal penalty term, which encourages this behaviour, contributes more significantly to the score of the negative classes than the nGWP term, which relies on relative pixel weighting.

\begin{table*}
\footnotesize
\begin{tabularx}{\linewidth}{@{}X|S[table-format=2.1]@{\hspace{0.5em}}S[table-format=2.1]@{\hspace{0.5em}}S[table-format=2.1]@{\hspace{0.5em}}S[table-format=2.1]@{\hspace{0.5em}}S[table-format=2.1]@{\hspace{0.5em}}S[table-format=2.1]@{\hspace{0.5em}}S[table-format=2.1]@{\hspace{0.5em}}S[table-format=2.1]@{\hspace{0.5em}}S[table-format=2.1]@{\hspace{0.5em}}S[table-format=2.1]@{\hspace{0.5em}}S[table-format=2.1]@{\hspace{0.5em}}S[table-format=2.1]@{\hspace{0.5em}}S[table-format=2.1]@{\hspace{0.5em}}S[table-format=2.1]@{\hspace{0.5em}}S[table-format=2.1]@{\hspace{0.5em}}S[table-format=2.1]@{\hspace{0.5em}}S[table-format=2.1]@{\hspace{0.5em}}S[table-format=2.1]@{\hspace{0.5em}}S[table-format=2.1]@{\hspace{0.5em}}S[table-format=2.1]@{\hspace{0.5em}}S[table-format=2.1]@{\hspace{0.5em}}|S[table-format=2.1]@{}}
\toprule
Method & {bg} & {aero} & {bike} & {bird} & {boat} & {bot.} & {bus} & {car} & {cat} & {chair} & {cow} & {tab.} & {dog} & {horse} & {mbk} & {per.} & {plant} & {sheep} & {sofa} & {train} & {tv} & {mean} \\
\midrule
\multicolumn{23}{@{}l}{\scriptsize \textit{Multi stage}} \\
\midrule
SEC$^\ast$ \cite{KolesnikovL16} & 82.4 & 62.9 & 26.4 & 61.6 & 27.6 & 38.1 & 66.6 & 62.7 & 75.2 & 22.1 & 53.5 & 28.3 & 65.8 & 57.8 & 62.3 & 52.5 & 32.5 & 62.6 & 32.1 & 45.4 & 45.3 & 50.7 \\
AdvErasing$^\ast$ \cite{WeiFLCZY17} & 83.4 & 71.1 & 30.5 & 72.9 & 41.6 & 55.9 & 63.1 & 60.2 & 74.0 & 18.0 & 66.5 & 32.4 & 71.7 & 56.3 & 64.8 & 52.4 & 37.4 & 69.1 & 31.4 & 58.9 & 43.9 & 55.0 \\
RDC$^\ast$ \cite{WeiXSJFH18} & 89.4 & 85.6 & 34.6 & 75.8 & 61.9 & 65.8 & 67.1 & 73.3 & 80.2 & 15.1 & 69.9 & 8.1 & 75.0 & 68.4 & 70.9 & 71.5 & 32.6 & 74.9 & 24.8 & 73.2 & 50.8 & 60.4 \\
FickleNet$^\ast$ \cite{LeeKLLY19} & 88.1 & 75.0 & 31.3 & 75.7 & 48.8 & 60.1 & 80.0 & 72.7 & 79.6 & 25.7 & 67.3 & 42.2 & 77.1 & 67.5 & 65.4 & 69.2 & 42.2 & 74.1 & 34.2 & 53.7 & 54.7 & 61.2 \\
AffinityNet~\cite{AhnK18} & 88.2 & 68.2 & 30.6 & 81.1 & 49.6 & 61.0 & 77.8 & 66.1 & 75.1 & 29.0 & 66.0 & 40.2 & 80.4 & 62.0 & 70.4 & 73.7 & 42.5 & 70.7 & 42.5 & 68.1 & 51.6 & 61.7 \\
FickleNet$^{\ast,\dagger}$ \cite{LeeKLLY19} & 89.5 & 76.6 & 32.6 & 74.6 & 51.5 & 71.1 & 83.4 & 74.4 & 83.6 & 24.1 & 73.4 & 47.4 & 78.2 & 74.0 & 68.8 & 73.2 & 47.8 & 79.9 & 37.0 & 57.3 & 64.6 & 64.9 \\
SSDD~\cite{Shimoda_2019_ICCV} & 89.0 & 62.5 & 28.9 & 83.7 & 52.9 & 59.5 & 77.6 & 73.7 & 87.0 & 34.0 & 83.7 & 47.6 & 84.1 & 77.0 & 73.9 & 69.6 & 29.8 & 84.0 & 43.2 & 68.0 & 53.4 & 64.9 \\
\midrule
\multicolumn{23}{@{}l}{\scriptsize \textit{Single stage}} \\
\midrule
TransferNet$^\ast$ \cite{HongOLH16} & 85.3 & 68.5 & 26.4 & 69.8 & 36.7 & 49.1 & 68.4 & 55.8 & 77.3 & 6.2 & 75.2 & 14.3 & 69.8 & 71.5 & 61.1 & 31.9 & 25.5 & 74.6 & 33.8 & 49.6 & 43.7 & 52.1 \\
CRF-RNN \cite{RoyT17} & 85.8 & 65.2 & 29.4 & 63.8 & 31.2 & 37.2 & 69.6 & 64.3 & 76.2 & 21.4 & 56.3 & 29.8 & 68.2 & 60.6 & 66.2 & 55.8 & 30.8 & 66.1 & 34.9 & 48.8 & 47.1 & 52.8 \\
WebCrawl$^\ast$ \cite{HongYKLH17} & 87.0 & 69.3 & 32.2 & 70.2 & 31.2 & 58.4& 73.6 & 68.5 & 76.5 & 26.8 & 63.8 & 29.1 & 73.5 & 69.5 & 66.5 & 70.4 & 46.8 & 72.1 & 27.3 & 57.4 & 50.2 & 58.1 \\
\midrule
Ours & 87.0 & 63.4 & 33.1 & 64.5 & 47.4 & 63.2 & 70.2 & 59.2 & 76.9 & 27.3 & 67.1 & 29.8 & 77.0 & 67.2 & 64.0 & 72.4 & 46.5 & 67.6 & 38.1 & 68.2 & 63.6 & 59.7 \\
Ours + CRF & 88.7 & 70.4 & 35.1 & 75.7 & 51.9 & 65.8 & 71.9 & 64.2 & 81.1 & 30.8 & 73.3 & 28.1 & 81.6 & 69.1 & 62.6 & 74.8 & 48.6 & 71.0 & 40.1 & 68.5 & 64.3 & 62.7  \\
\bottomrule
\multicolumn{23}{@{}l}{\scriptsize Methods marked with $(^\ast)$ use saliency detectors or additional data, or both (see Sec. 2). $(^\dagger)$ denotes a ResNet-101 backbone.} \\
\end{tabularx}
\caption{Per-class IoU (\%) comparison on Pascal VOC 2012, validation set.}
\label{table:main_result_val}
\end{table*}

\begin{table*}
\footnotesize
\begin{tabularx}{\linewidth}{@{}X|S[table-format=2.1]@{\hspace{0.5em}}S[table-format=2.1]@{\hspace{0.5em}}S[table-format=2.1]@{\hspace{0.5em}}S[table-format=2.1]@{\hspace{0.5em}}S[table-format=2.1]@{\hspace{0.5em}}S[table-format=2.1]@{\hspace{0.5em}}S[table-format=2.1]@{\hspace{0.5em}}S[table-format=2.1]@{\hspace{0.5em}}S[table-format=2.1]@{\hspace{0.5em}}S[table-format=2.1]@{\hspace{0.5em}}S[table-format=2.1]@{\hspace{0.5em}}S[table-format=2.1]@{\hspace{0.5em}}S[table-format=2.1]@{\hspace{0.5em}}S[table-format=2.1]@{\hspace{0.5em}}S[table-format=2.1]@{\hspace{0.5em}}S[table-format=2.1]@{\hspace{0.5em}}S[table-format=2.1]@{\hspace{0.5em}}S[table-format=2.1]@{\hspace{0.5em}}S[table-format=2.1]@{\hspace{0.5em}}S[table-format=2.1]@{\hspace{0.5em}}S[table-format=2.1]@{\hspace{0.5em}}|S[table-format=2.1]@{}}
\toprule
Method & {bg} & {aero} & {bike} & {bird} & {boat} & {bot.} & {bus} & {car} & {cat} & {chair} & {cow} & {tab.} & {dog} & {horse} & {mbk} & {per.} & {plant} & {sheep} & {sofa} & {train} & {tv} & {mean} \\
\midrule
\multicolumn{23}{@{}l}{\scriptsize \textit{Multi stage}} \\
\midrule
SEC$^\ast$ \cite{KolesnikovL16} & 83.5 & 56.4 & 28.5 & 64.1 & 23.6 & 46.5 & 70.6 & 58.5 & 71.3 & 23.2 & 54.0 & 28.0 & 68.1 & 62.1 & 70.0 & 55.0 & 38.4 & 58.0 & 39.9 & 38.4 & 48.3 & 51.7 \\
FickleNet$^\ast$ \cite{LeeKLLY19} & 88.5 & 73.7 & 32.4 & 72.0 & 38.0 & 62.8 & 77.4 & 74.4 & 78.6 & 22.3 & 67.5 & 50.2 & 74.5 & 72.1 & 77.3 & 68.8 & 52.5 & 74.8 & 41.5 & 45.5 & 55.4 & 61.9 \\
AffinityNet~\cite{AhnK18} & 89.1 & 70.6 & 31.6 & 77.2 & 42.2 & 68.9 & 79.1 & 66.5 & 74.9 & 29.6 & 68.7 & 56.1 & 82.1 & 64.8 & 78.6 & 73.5 & 50.8 & 70.7 & 47.7 & 63.9 & 51.1 & 63.7 \\
FickleNet$^{\ast,\dagger}$ \cite{LeeKLLY19} & 89.8 & 78.3 & 34.1 & 73.4 & 41.2 & 67.2 & 81.0 & 77.3 & 81.2 & 29.1 & 72.4 & 47.2 & 76.8 & 76.5 & 76.1 & 72.9 & 56.5 & 82.9 & 43.6 & 48.7 & 64.7 & 65.3 \\
SSDD \cite{Shimoda_2019_ICCV} & 89.5 & 71.8 & 31.4 & 79.3 & 47.3 & 64.2 & 79.9 & 74.6 & 84.9 & 30.8 & 73.5 & 58.2 & 82.7 & 73.4 & 76.4 & 69.9 & 37.4 & 80.5 & 54.5 & 65.7 & 50.3 & 65.5 \\
\midrule
\multicolumn{23}{@{}l}{\scriptsize \textit{Single stage}} \\
\midrule
TransferNet$^\ast$~\cite{HongOLH16} & 85.7 & 70.1 & 27.8 & 73.7 & 37.3 & 44.8 & 71.4 & 53.8 & 73.0 & 6.7 & 62.9 & 12.4 & 68.4 & 73.7 & 65.9 & 27.9 & 23.5 & 72.3 & 38.9 & 45.9 & 39.2 & 51.2 \\
CRF-RNN \cite{RoyT17} & 85.7 & 58.8 & 30.5 & 67.6 & 24.7 & 44.7 & 74.8 & 61.8 & 73.7 & 22.9 & 57.4 & 27.5 & 71.3 & 64.8 & 72.4 & 57.3 & 37.3 & 60.4 & 42.8 & 42.2 & 50.6 & 53.7 \\
WebCrawl$^\ast$ \cite{HongYKLH17} & 87.2 & 63.9 & 32.8 & 72.4 & 26.7 & 64.0 & 72.1 & 70.5 & 77.8 & 23.9 & 63.6 & 32.1 & 77.2 & 75.3 & 76.2 & 71.5 & 45.0 & 68.8 & 35.5 & 46.2 & 49.3 & 58.7 \\
\midrule
Ours & 87.4 & 63.6 & 34.7 & 59.9 & 40.1 & 63.3 & 70.2 & 56.5 & 71.4 & 29.0 & 71.0 & 38.3 & 76.7 & 73.2 & 70.5 & 71.6 & 55.0 & 66.3 & 47.0 & 63.5 & 60.3 & 60.5 \\
Ours + CRF & 89.2 & 73.4 & 37.3 & 68.3 & 45.8 & 68.0 & 72.7 & 64.1 & 74.1 & 32.9 & 74.9 & 39.2 & 81.3 & 74.6 & 72.6 & 75.4 & 58.1 & 71.0 & 48.7 & 67.7 & 60.1 & 64.3 \\
\bottomrule
\multicolumn{23}{@{}l}{\scriptsize Methods marked with $(^\ast)$ use saliency detectors or additional data, or both (see Sec. 2). $(^\dagger)$ denotes a ResNet-101 backbone.}  \\
\end{tabularx}

\caption{Per-class IoU (\%) comparison on Pascal VOC 2012, test set.}
\label{table:main_result_test}
\end{table*}

\item \textbf{Negative class debiasing}.
Negative classes dominate in the label set of Pascal VOC, \ie each image sample depicts only few categories.
The gradient of the loss function from \cref{eq:supp_cls_loss} vanishes for positive classes with $y_{c} \rightarrow \infty$ and negative classes with $y_{c} \rightarrow -\infty$.
However, in our preliminary experiments with GAP-CAM, and later with nGWP with $\epsilon = 0$ in \cref{eq:supp_gwp}, we observed that further iterations continued to decrease the scores for the negative classes in regions in which the loss is near-saturated, while the $y_{c}$ of positive classes increased only marginally.
This may indicate a strong inductive bias towards negative classes, which might be undesirable for real-world deployment.

Assuming $\epsilon > 0$ and $\sum_{i',j'} m_{c,i',j'} \neq 0$, we observe that,
\begin{equation}
\begin{aligned}
\frac{\sum_{i,j} m_{c,i,j} y_{c,i,j}}{\epsilon + \sum_{i',j'} m_{c,i',j'}} > \frac{\sum_{i,j} m_{c,i,j} y_{c,i,j}}{\sum_{i',j'} m_{c,i',j'}}
\end{aligned}
\label{eq:supp_ineq}
\end{equation}
when $y_{c,:,:} < 0$, which is the case for saturated negative classes.
Recall further that the use of the constant background score (fixed to $1$) implies that $m_{c,i,j} \rightarrow 0$ as $y_{c,i,j} \rightarrow -\infty$.
Since the RHS is the convex combination of $y_{c,i,j}$'s, its minimum is $\min(\{y_{c,i,j}\}_{i,j})$.
Therefore, the RHS is unbounded from below as $y_{c,i,j} \rightarrow -\infty$, hence the $y_{c,i,j}$ keep getting pushed down for negative classes.
By contrast, the LHS has a defined limit of $0$ as $m_{c,i,j} \rightarrow 0$ (see \cref{eq:supp_limit_zero}), which is undesirable as the score for a negative class.
This is because a finite (\ie larger) $m_{c,i,j}$ yields a negative, thus smaller value of the classification score, which we are trying to minimise in this case.
Therefore, $\epsilon>0$ will encourage SGD to pull the negative scores away from the saturation areas by pushing the $\sum_{i,j} m_{c,i,j}$ away from zero.
\end{enumerate}

In summary, for negative classes $\epsilon$ improves numerical stability and emphasises the focal penalty while leveraging nGWP to alleviate the negative class bias.
For positive classes, the effect of $\epsilon$ is negligible, since $\epsilon \ll \sum_{i,j} m_{c,i,j}$ in this case.
We set $\epsilon = 1$ in all our experiments.

\subsection{Segmentation loss}
For the segmentation loss \wrt the pseudo ground truth, we use a \textit{weighted} cross-entropy defined for each pixel site $(i, j)$ as
\begin{equation}
\begin{aligned}
\mathcal{L}_{\text{seg},i,j} = - q_c \log m_{c,i,j},
\end{aligned}
\end{equation}
where $c$ is the label in the pseudo ground-truth mask.
The balancing class weight $q_c$ accounts for the fact that the pseudo ground truth may contain a different amount of pixel supervision for each class:
\begin{equation}
\begin{aligned}
q_c = \frac{M_{b,\text{total}} - M_{b,c}}{1 + M_{b,\text{total}}},
\end{aligned}
\end{equation}
where $M_{b,\text{total}}$ and $M_{b,c}$ are the total and class-specific number of pixels for supervision in the pseudo ground-truth mask, respectively;
$b$ indexes the sample in a batch;
and $1$ in the denominator serves for numerical stability.
The aggregated segmentation loss $\mathcal{L}_\text{seg}$ is a weighted mean over the samples in a batch, \ie
\begin{equation}
\begin{aligned}
\mathcal{L}_\text{seg} = \frac{1}{\sum_{b^\prime} M_{b^\prime,\text{total}}} \frac{1}{hw} \sum_b M_{b,\text{total}} \sum_{i,j} \mathcal{L}_{\text{seg},i,j}.
\end{aligned}
\end{equation}

\section{Quantitative Analysis}
\label{sec:supp_qaunt}
We list the per-class segmentation accuracy on Pascal VOC validation and training in \cref{table:main_result_val,table:main_result_test}.
We first observe that none of previous methods, including the state of the art \cite{AhnK18,LeeKLLY19,Shimoda_2019_ICCV}, outperforms other pipelines on \emph{all} class categories.
For example FickleNet \cite{LeeKLLY19}, based on a ResNet-101 backbone, reaches top segmentation accuracy only for classes ``bottle'', ``bus'', ``car'', and ``tv''.
SSDD \cite{Shimoda_2019_ICCV} has the highest mean IoU, but is inferior to other methods on 10 out of 21 classes.
Our single-stage method compares favourably even to the multi-stage approaches that rely on saliency supervision or additional data.
For example, we improve over the more complex AffinityNet~\cite{AhnK18} that trains a deep network to predict pixel-level affinity distance from CAMs and then further trains a segmentation network in a (pseudo) fully supervised regime.
The best single-stage method from previous work, CRF-RNN \cite{RoyT17}, trained using only the image-level annotation we consider in this work, reaches $52.8\%$ and $53.7\%$ IoU on the validation and test sets.
We substantially boost this result, by $9.9\%$ and $10.6\%$ points, respectively, and attain new a state-of-the-art mask accuracy overall on classes ``bike'', ``person'', and ``plant''.

\section{Ablation Study: PAMR Iterations}

We empirically verify the number of iterations used in PAMR, which we set to $10$ in our main ablation study.
\cref{table:pamr_iter} reports the results with higher and fewer number of iterations.
We observe that using only few PAMR iterations decreases the mask quality.
On the other hand, the benefits of PAMR diminishes if we increase the number of iterations further.
$10$ iterations appears to strike a good balance between the computational expense and the obtained segmentation accuracy.

\begin{figure*}[t]
\captionsetup[subfigure]{labelformat=empty}
\centering
\begin{subfigure}{0.137\linewidth}
\caption{Prediction}
\includegraphics[width=\linewidth]{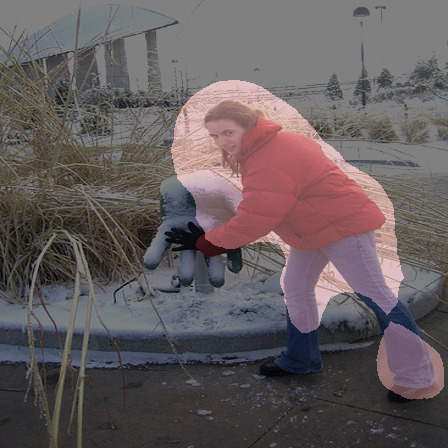}\end{subfigure}
\begin{subfigure}{0.137\linewidth}
\caption{Iteration 1}
\includegraphics[width=\linewidth]{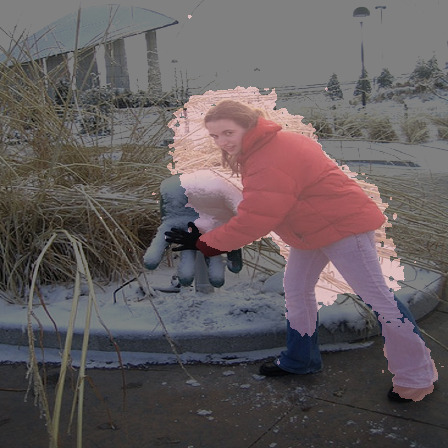}\end{subfigure}
\begin{subfigure}{0.137\linewidth}
\caption{Iteration 3}
\includegraphics[width=\linewidth]{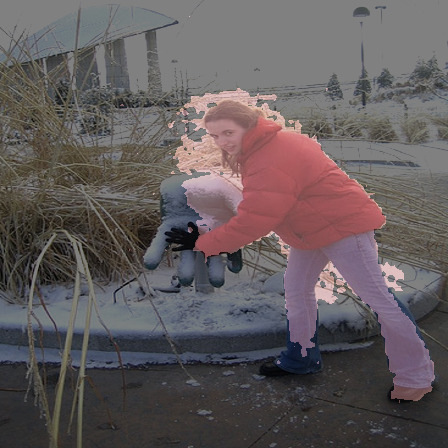}\end{subfigure}
\begin{subfigure}{0.137\linewidth}
\caption{Iteration 5}
\includegraphics[width=\linewidth]{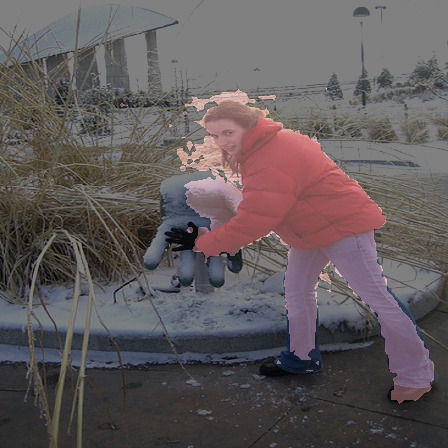}\end{subfigure}
\begin{subfigure}{0.137\linewidth}
\caption{Iteration 7}
\includegraphics[width=\linewidth]{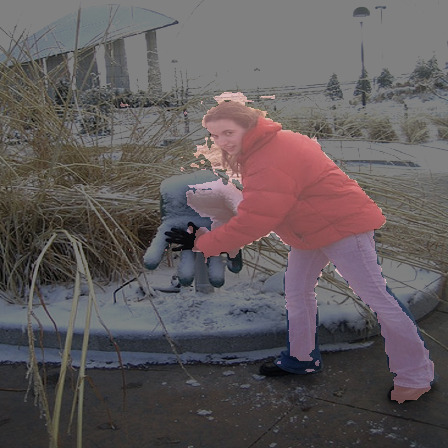}\end{subfigure}
\begin{subfigure}{0.137\linewidth}
\caption{Iteration 10}
\includegraphics[width=\linewidth]{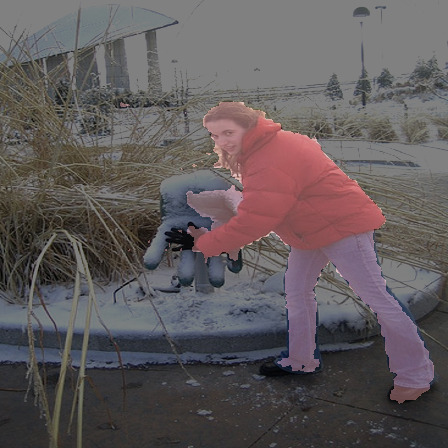}\end{subfigure}
\begin{subfigure}{0.137\linewidth}
\caption{Ground Truth}
\includegraphics[width=\linewidth]{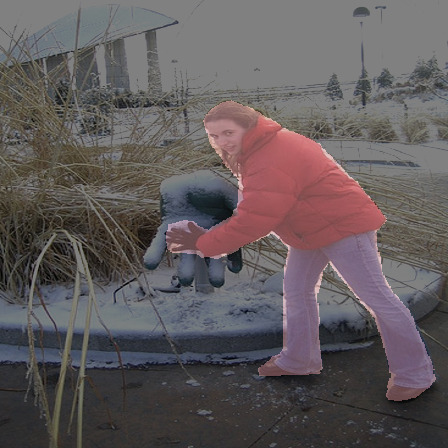}\end{subfigure}

\vspace{1mm}
\begin{subfigure}{0.137\linewidth}\includegraphics[width=\linewidth]{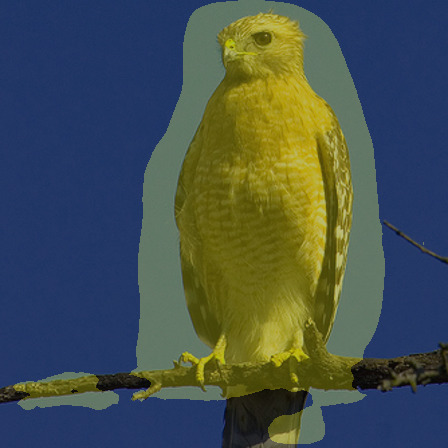}\end{subfigure}
\begin{subfigure}{0.137\linewidth}\includegraphics[width=\linewidth]{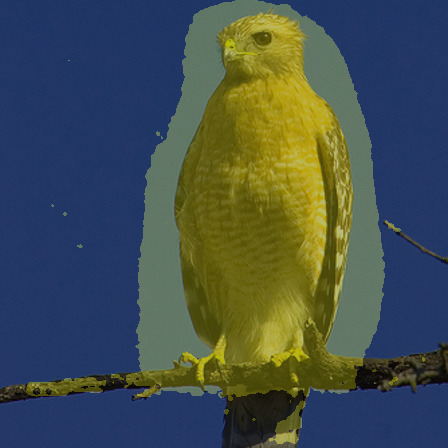}\end{subfigure}
\begin{subfigure}{0.137\linewidth}\includegraphics[width=\linewidth]{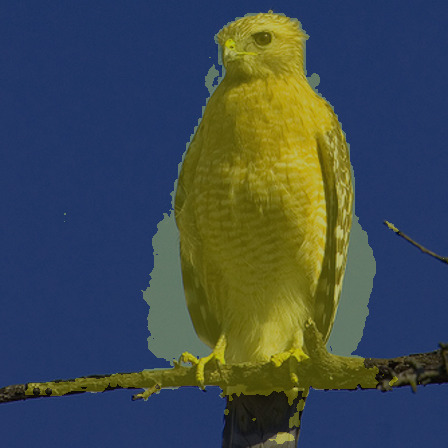}\end{subfigure}
\begin{subfigure}{0.137\linewidth}\includegraphics[width=\linewidth]{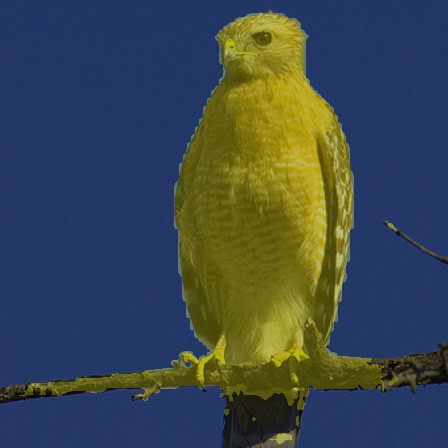}\end{subfigure}
\begin{subfigure}{0.137\linewidth}\includegraphics[width=\linewidth]{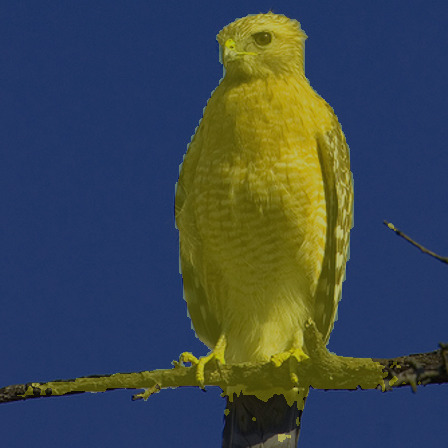}\end{subfigure}
\begin{subfigure}{0.137\linewidth}\includegraphics[width=\linewidth]{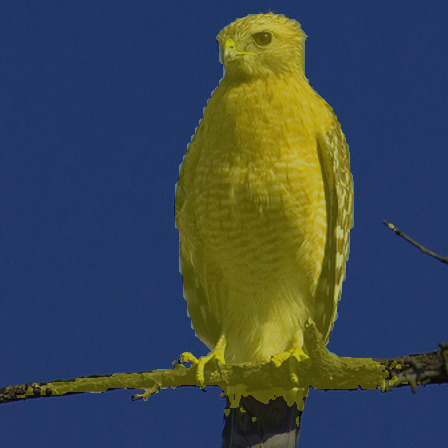}\end{subfigure}
\begin{subfigure}{0.137\linewidth}\includegraphics[width=\linewidth]{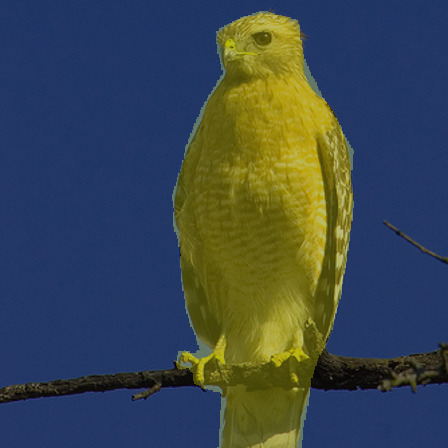}\end{subfigure}

\vspace{1mm}
\begin{subfigure}{0.137\linewidth}\includegraphics[width=\linewidth]{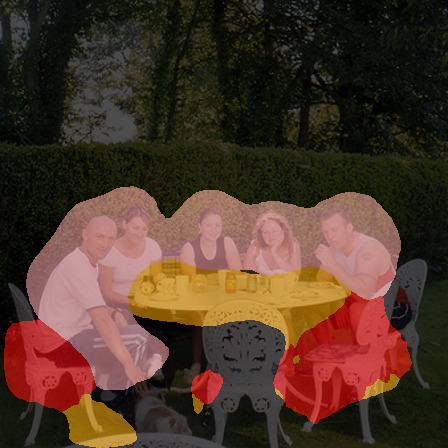}\end{subfigure}
\begin{subfigure}{0.137\linewidth}\includegraphics[width=\linewidth]{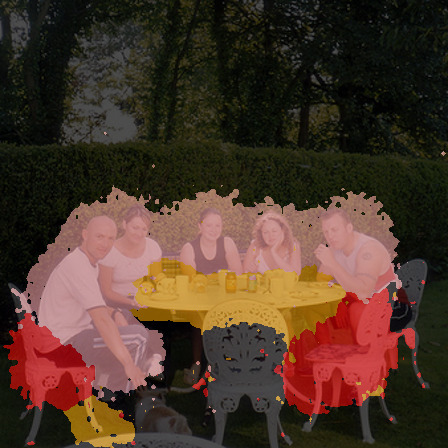}\end{subfigure}
\begin{subfigure}{0.137\linewidth}\includegraphics[width=\linewidth]{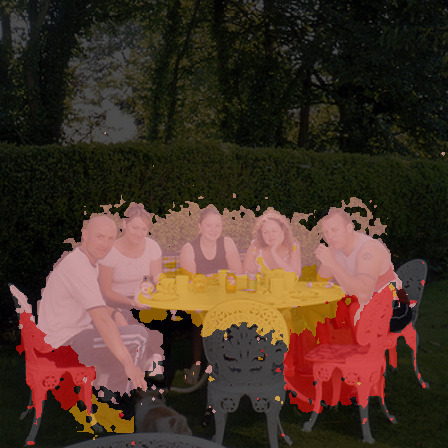}\end{subfigure}
\begin{subfigure}{0.137\linewidth}\includegraphics[width=\linewidth]{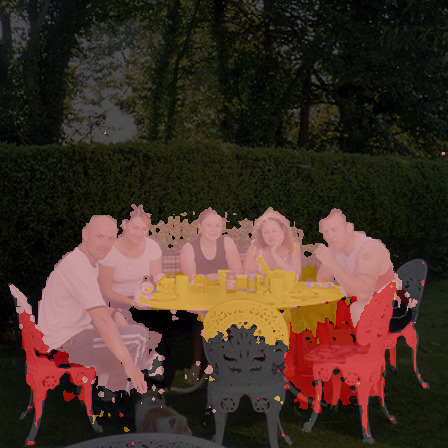}\end{subfigure}
\begin{subfigure}{0.137\linewidth}\includegraphics[width=\linewidth]{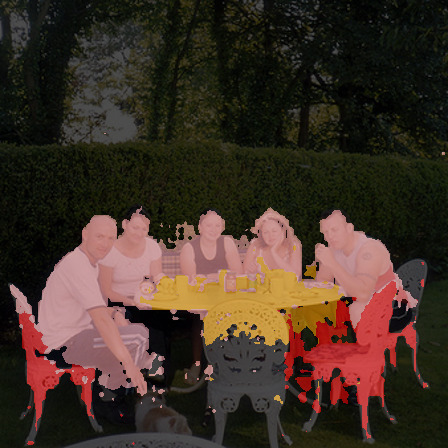}\end{subfigure}
\begin{subfigure}{0.137\linewidth}\includegraphics[width=\linewidth]{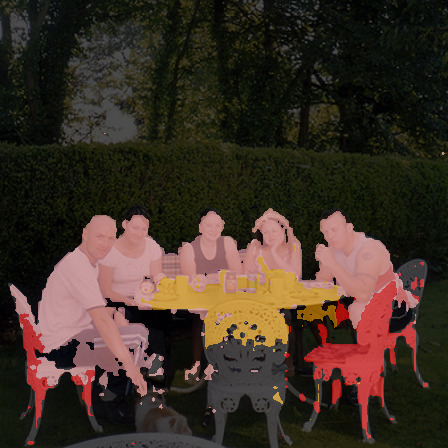}\end{subfigure}
\begin{subfigure}{0.137\linewidth}\includegraphics[width=\linewidth]{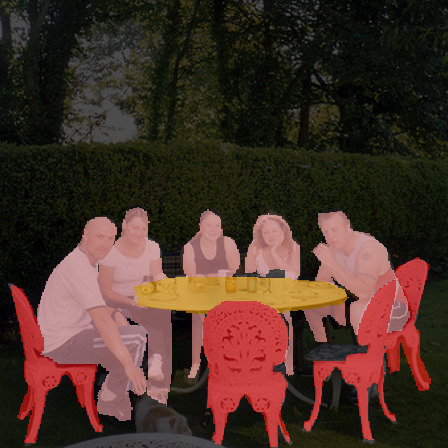}\end{subfigure}

\vspace{1mm}
\begin{subfigure}{0.137\linewidth}\includegraphics[width=\linewidth]{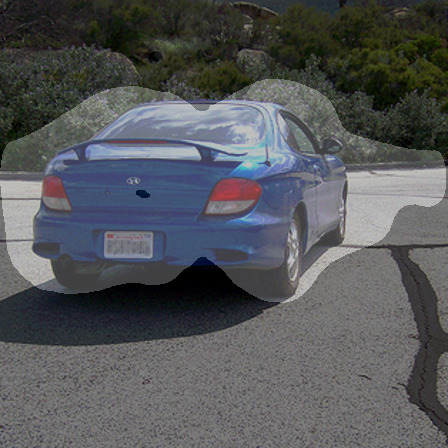}\end{subfigure}
\begin{subfigure}{0.137\linewidth}\includegraphics[width=\linewidth]{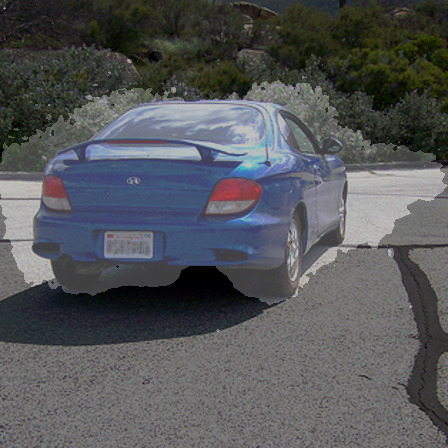}\end{subfigure}
\begin{subfigure}{0.137\linewidth}\includegraphics[width=\linewidth]{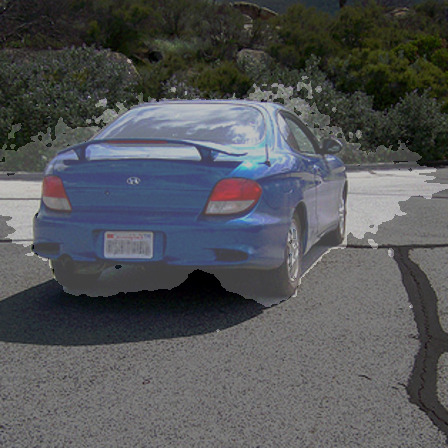}\end{subfigure}
\begin{subfigure}{0.137\linewidth}\includegraphics[width=\linewidth]{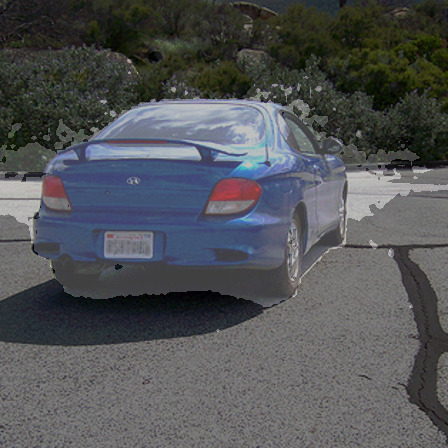}\end{subfigure}
\begin{subfigure}{0.137\linewidth}\includegraphics[width=\linewidth]{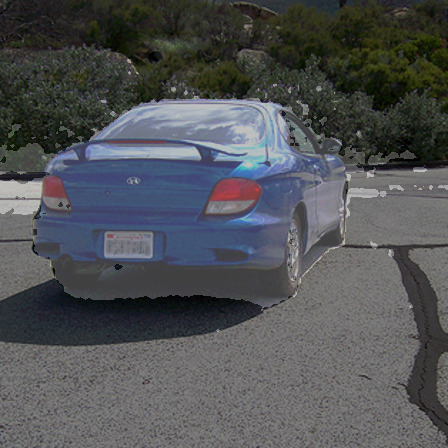}\end{subfigure}
\begin{subfigure}{0.137\linewidth}\includegraphics[width=\linewidth]{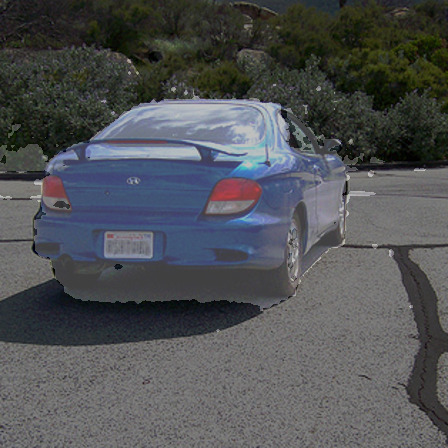}\end{subfigure}
\begin{subfigure}{0.137\linewidth}\includegraphics[width=\linewidth]{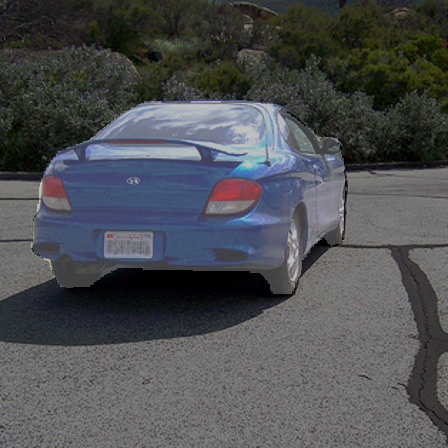}\end{subfigure}

\vspace{1mm}
\begin{subfigure}{0.137\linewidth}\includegraphics[width=\linewidth]{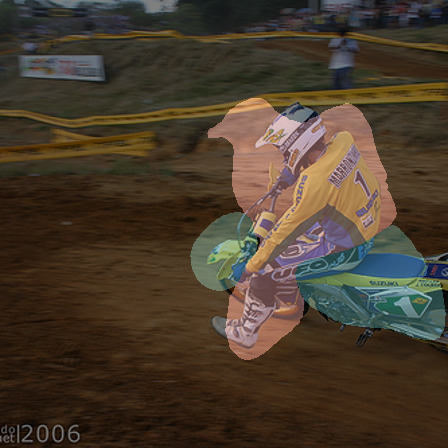}\end{subfigure}
\begin{subfigure}{0.137\linewidth}\includegraphics[width=\linewidth]{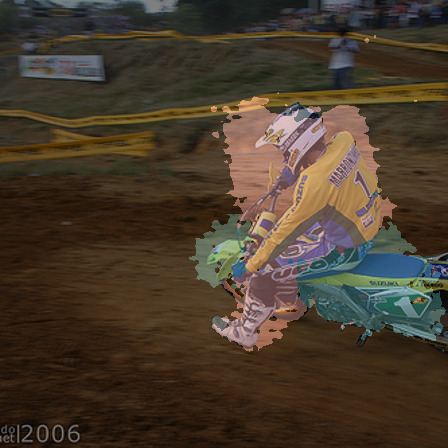}\end{subfigure}
\begin{subfigure}{0.137\linewidth}\includegraphics[width=\linewidth]{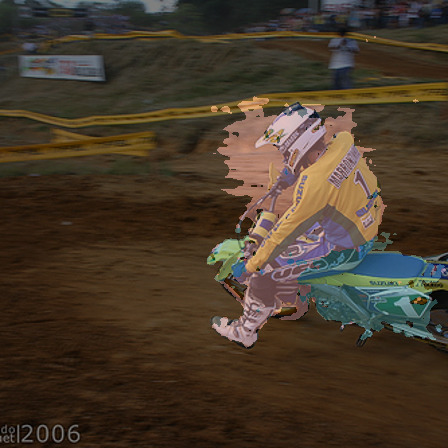}\end{subfigure}
\begin{subfigure}{0.137\linewidth}\includegraphics[width=\linewidth]{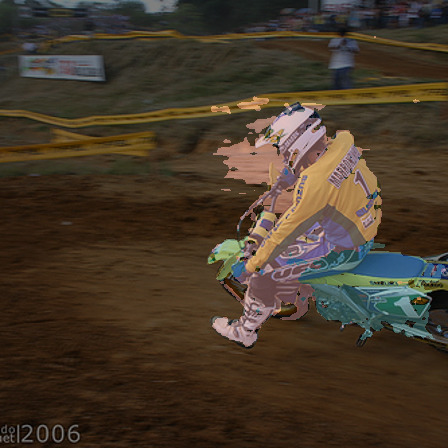}\end{subfigure}
\begin{subfigure}{0.137\linewidth}\includegraphics[width=\linewidth]{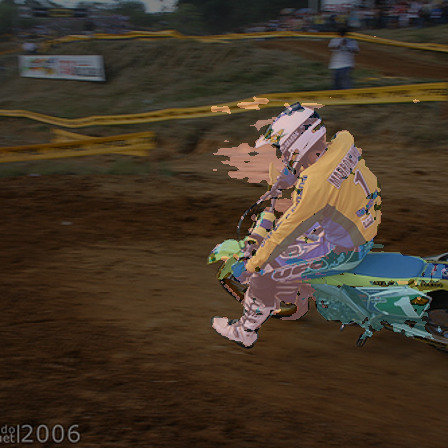}\end{subfigure}
\begin{subfigure}{0.137\linewidth}\includegraphics[width=\linewidth]{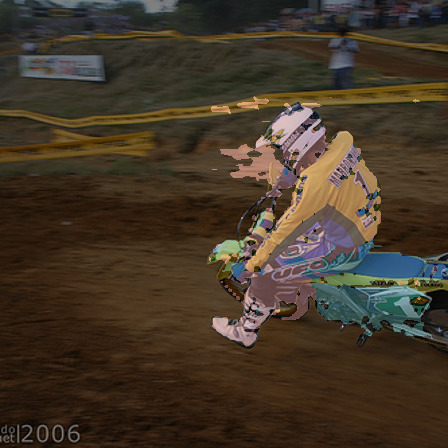}\end{subfigure}
\begin{subfigure}{0.137\linewidth}\includegraphics[width=\linewidth]{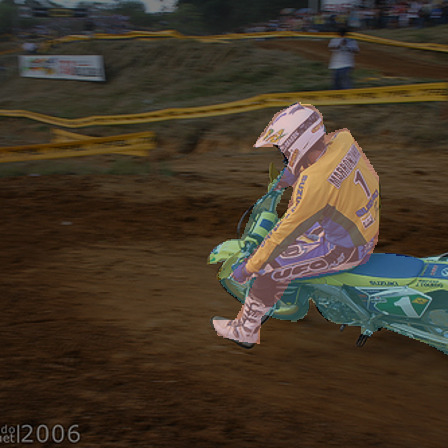}\end{subfigure}

\vspace{1mm}
\begin{subfigure}{0.137\linewidth}\includegraphics[width=\linewidth]{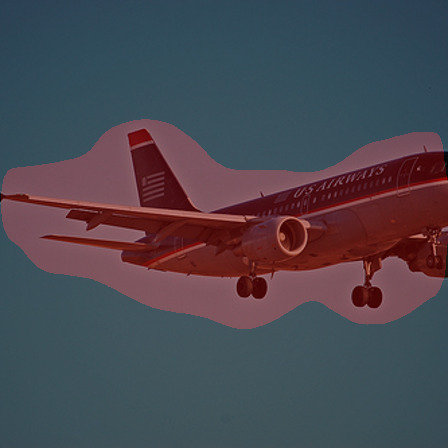}\end{subfigure}
\begin{subfigure}{0.137\linewidth}\includegraphics[width=\linewidth]{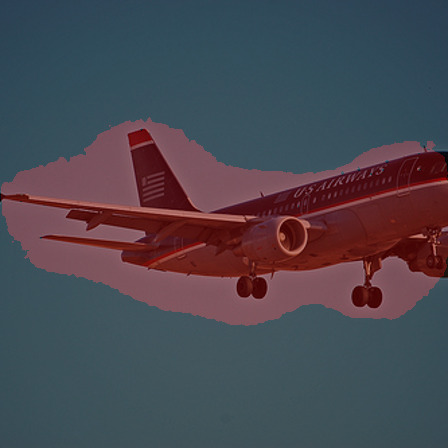}\end{subfigure}
\begin{subfigure}{0.137\linewidth}\includegraphics[width=\linewidth]{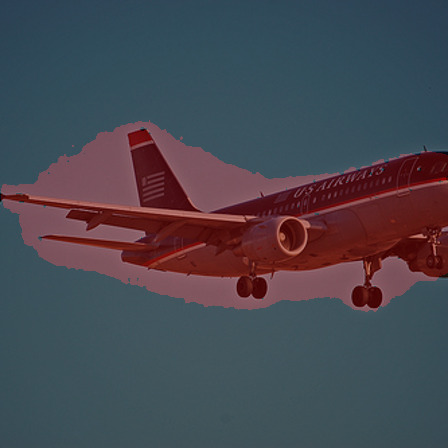}\end{subfigure}
\begin{subfigure}{0.137\linewidth}\includegraphics[width=\linewidth]{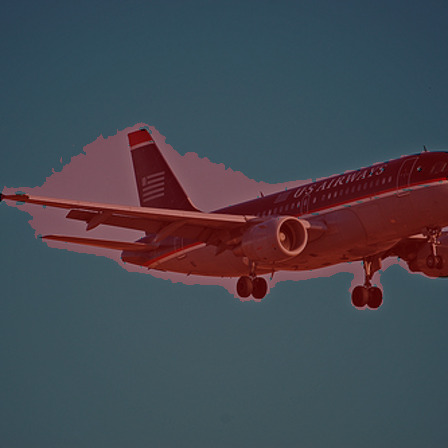}\end{subfigure}
\begin{subfigure}{0.137\linewidth}\includegraphics[width=\linewidth]{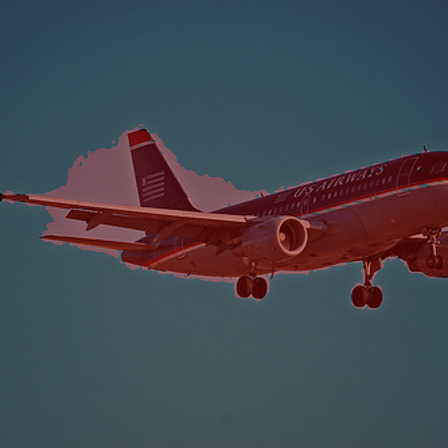}\end{subfigure}
\begin{subfigure}{0.137\linewidth}\includegraphics[width=\linewidth]{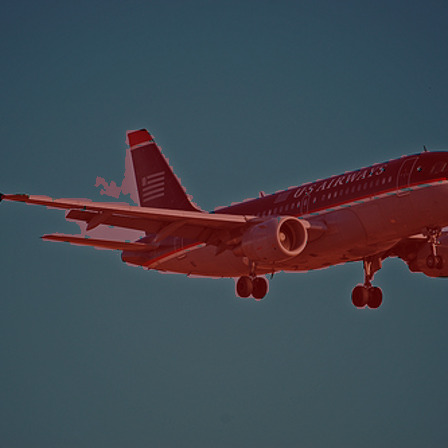}\end{subfigure}
\begin{subfigure}{0.137\linewidth}\includegraphics[width=\linewidth]{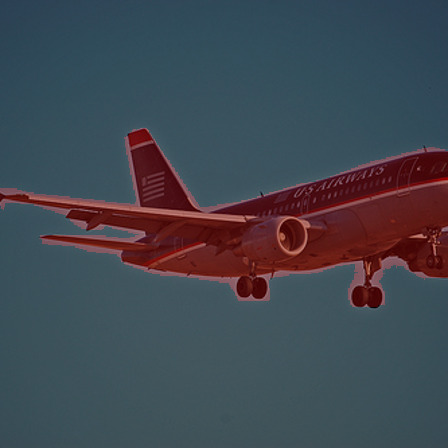}\end{subfigure}

\caption{\textbf{Visualisation of PAMR iterations.} The initial model predictions suffer from local inconsistency: mask boundaries do not align with available visual cues. Our PAMR module iteratively revises the masks to alleviate this problem. Our model uses the mask from the last iteration for self-supervision.}
\label{fig:pamr_iter}
\vspace{-0.5em}
\end{figure*}

Additionally, we visualise semantic masks produced at intermediate iterations from our PAMR module in \cref{fig:pamr_iter}.
The initial masks produced by our segmentation model in the early stages of training exhibit coarse boundaries.
PAMR mitigates this shortcoming by virtue of exploiting visual cues with pixel-adaptive convolutions.
Our model then uses the revised masks to generate pseudo ground-truth for self-supervised segmentation.

\begin{table}
\footnotesize
\begin{tabularx}{\linewidth}{@{}XS[table-format=2.1]@{\hspace{2em}}S[table-format=2.1]@{}}
\toprule
Number of iterations & {IoU} & {IoU w/ CRF} \\
\midrule
5 & 59.1 & 59.0 \\
10 & \bfseries 59.4 & \bfseries 62.2 \\
15 & 59.0 & 61.6 \\
\bottomrule
\end{tabularx}
\caption{\textbf{Effect of the iteration number in PAMR.} We train our model with the iteration number in the PAMR module fixed to a pre-defined value. We report the IoU (\%) with and without CRF refinement on Pascal VOC validation.}
\label{table:pamr_iter}
\end{table}

\section{Pseudo Labels}
\label{sec:pseudo_labels}

The last-stage training of a segmentation network is \emph{agnostic} to the process of pseudo-label generation;
it is the quality of the pseudo labels and the ease of obtaining them that matters.

\begin{table}
\footnotesize
\begin{tabularx}{\linewidth}{@{}X@{\hspace{0.5em}}lcS[table-format=2.1]@{\hspace{1em}}S[table-format=2.1]@{}}
\toprule
Method & Backbone & Supervision & {val} & {test} \\
\midrule
\multicolumn{5}{@{}l}{\scriptsize \textit{Fully supervised}} \\
\midrule
WideResNet38~\cite{wu2019wider} & & $\mathcal{F}$ & 80.8 & 82.5 \\
DeepLabv3+~\cite{deeplabv3plus2018} & Xception-65 \cite{chollet17} & $\mathcal{F}$ & {--} & 87.8 \\
\midrule
\multicolumn{5}{@{}l}{\scriptsize \textit{Multi stage + Saliency}} \\
\midrule
FickleNet~\cite{LeeKLLY19} & ResNet-101 & $\mathcal{S}$ & 64.9 & 65.3 \\
\multirow{2}{*}{Frame-to-Frame~\cite{Lee_2019_ICCV}} & VGG-16 & \multirow{2}{*}{$\mathcal{S}, \mathcal{D}$} & 63.9 & 65.0 \\
 & ResNet-101 & & 66.5 & 67.4 \\
\midrule
\multicolumn{5}{@{}l}{\scriptsize \textit{Single stage + Saliency}} \\
\midrule
JointSaliency~\cite{Zeng_2019_ICCV} & \scriptsize{DenseNet-169} \cite{HuangLMW17} & $\mathcal{S}, \mathcal{D}$ & 63.3 & 64.3 \\
\midrule
\multicolumn{5}{@{}l}{\scriptsize \textit{Multi stage}} \\
\midrule
AffinityNet~\cite{AhnK18} & WideResNet-38 & $\mathcal{I}$ & 61.7 & 63.7 \\
IRN~\cite{AhnCK19} & ResNet-50 & $\mathcal{I}$ & 63.5 & 64.8 \\
SSDD~\cite{Shimoda_2019_ICCV} & WideResNet-38 & $\mathcal{I}$ & 64.9 & 65.5 \\
\midrule
\multicolumn{5}{@{}l}{\scriptsize \textit{Two stage}} \\
\midrule
Ours + DeepLabv3+ & ResNet-101 & $\mathcal{I}$ & 65.7 & 66.6 \\
Ours + DeepLabv3+ & Xception-65 & $\mathcal{I}$ & 66.8 & 67.3\\
\bottomrule
\end{tabularx}
\caption{\textbf{Mean IoU (\%) on Pascal VOC validation and test sets.}
We train DeepLabv3+ in a fully supervised regime on pseudo ground truth obtained from our method (with CRF refinement).
Under equivalent level of supervision, our two-stage approach outperforms the previous state of the art, trained in three or more stages
and performs on par with other multi-stage frameworks relying on additional data and saliency detection \cite{Lee_2019_ICCV}.
}
\label{table:finetuning}
\end{table}

Although we intentionally omitted the common practice of training a standalone network on our pseudo labels,
we show that, in fact, we can achieve state-of-the-art results in a multi-stage setting as well.

We use our pseudo labels on the train split of Pascal VOC (see \cref{table:seg_training}) and train a segmentation model DeepLabv3+~\cite{deeplabv3plus2018} in a fully supervised fashion.
\cref{table:finetuning} summarises the results.
We observe that the resulting simple two-stage pipeline outperforms other multi-stage frameworks under the same image-level supervision.
Remarkably, our method even attains the mask accuracy of Frame-to-Frame~\cite{Lee_2019_ICCV}, which not only utilises saliency detectors, but also relies on additional data (15K extra) and sophisticated network models (\eg PWC-Net [46], FickleNet \cite{LeeKLLY19}).

\begin{table}
\small
\begin{tabularx}{\linewidth}{@{}XS[table-format=2.1]@{\hspace{1.5em}}S[table-format=2.1]@{\hspace{1em}}c@{\hspace{1em}}S[table-format=2.1]@{\hspace{1em}}S[table-format=2.1]@{}}
\toprule
& \multicolumn{2}{@{}c@{}}{Baseline (CAM)} & & \multicolumn{2}{@{}c@{}}{Ours} \\ \cmidrule(lr){2-3} \cmidrule{5-6}
Backbone & {w/o CRF} & {+ CRF} & & {w/o CRF} & {+ CRF} \\
\midrule
VGG16 & 41.2 & 38.0 & & 55.9 & 56.6 \\
ResNet-50 & 43.7 & 43.5 & & 60.4 & 64.1 \\
ResNet-101 & 46.2 & 45.2 & & 62.9 & 66.2 \\
WideResNet-38 & 44.9 & 45.2 & & 63.1 & 65.8 \\
\midrule
Mean & 44.0 & 43.0 & & 60.6 & 63.2 \\
\bottomrule
\end{tabularx}
\caption{\textbf{Segmentation quality (IoU, \%) on Pascal VOC validation.} We use ground-truth image-level labels to remove false positive classes from the masks to decouple the segmentation quality from the accuracy of the classifier.}
\label{table:backbones}
\end{table}

\section{Exchanging Backbones}
\label{sec:backbones}

Here we confirm that our segmentation method generalises well to other backbone architectures.
We choose VGG16 \cite{SimonyanZ14a}, ResNet-50 and ResNet-101 \cite{HeZRS16} -- all widely used network architectures -- as a drop-in alternative to WideResNet-38 \cite{wu2019wider}, which we use for all other experiments.
We train these models on $448\times448$ image crops using the same data augmentation as before.
We use multi-scale inference with image sides varying by factors $1.0$, $0.75$, $1.25$, $1.5$.
These scales are slightly different from the ones we used in the main experiments, which we found to slightly improve the IoU on average.
We re-evaluate our WideResNet-38 on the same scales to make the results from different backbones compatible.
Motivated to measure the segmentation accuracy alone, we report validation IoU on the masks with the false positives removed using ground-truth labels.
\cref{table:backbones} summarises the results.

The results show a clear improvement over the CAM baseline for all backbones: the average improvement without CRF post-processing is $16.6\%$ IoU and $20.2\%$ IoU with CRF refinement.

{\small
\bibliographystylesupp{ieee_fullname}
\bibliographysupp{egbib_supp}
}

\end{document}